\documentclass[conference]{IEEEtran}
\PassOptionsToPackage{capitalise,noabbrev,nameinlink}{cleveref}
\usepackage{times,latexsym}
\usepackage{graphicx} \graphicspath{{figures/}}
\usepackage{amsmath,amssymb,amsfonts,mathabx,mathtools,amsthm,nicefrac}
\usepackage[linesnumbered,ruled,vlined]{algorithm2e}
\usepackage{acronym}
\usepackage[bookmarks=true]{hyperref}
\usepackage{balance}
\usepackage{xspace}
\usepackage{setspace}
\usepackage[skip=3pt,font=small]{subcaption}
\usepackage[skip=3pt,font=small]{caption}
\usepackage[dvipsnames,svgnames,x11names,table]{xcolor}
\usepackage[capitalise,noabbrev,nameinlink]{cleveref}
\usepackage{booktabs,tabularx,colortbl,multirow,multicol,array,makecell,tabularray}
\let\labelindent\relax\usepackage{enumitem}
\usepackage[misc]{ifsym}
\usepackage{pifont}
\usepackage{tikz}
\usepackage{siunitx}
\usepackage[numbers]{natbib}
\usepackage{placeins}
\usepackage{dblfloatfix}
\usepackage[flushleft]{threeparttable}

\makeatletter
\DeclareRobustCommand\onedot{\futurelet\@let@token\@onedot}
\def\@onedot{\ifx\@let@token.\else.\null\fi\xspace}
\def\eg{\emph{e.g}\onedot}

\makeatother

\newcommand{\model}{\textbf{\texttt{OmniClone}}\xspace}
\newcommand{\bench}{\textbf{\texttt{OmniBench}}\xspace}
\newcommand{\groot}{{GR00T}\xspace}
\acrodef{dof}[DoF]{Degree-of-Freedom}
\acrodef{llm}[LLM]{Large Language Model}
\acrodef{rl}[RL]{Reinforcement Learning}
\acrodef{mpc}[MPC]{Model Predictive Control}
\acrodef{mlp}[MLPs]{Multi-layer Perceptrons}
\acrodef{slam}[SLAM]{Simultaneous Localization And Mapping}
\acrodef{mdp}[MDP]{Markov Decision Process}
\acrodef{ppo}[PPO]{Proximal policy optimization}
\acrodef{imu}[IMUs]{Inertial Measurement Units}
\acrodef{ood}[OOD]{Out-of-Distribution}
\acrodef{mocap}[MoCap]{Motion Capture}
\acrodef{sr}[SR]{Success Rate}
\acrodef{vla}[VLA]{Vision-Language-Action}
\acrodef{mpjpe}[MPJPE]{Mean Per-Joint Position Error}
\acrodef{sota}[SOTA]{state-of-the-art}
\acrodef{fk}[FK]{Forward kinematics}
\acrodef{ffn}[FFN]{Feed-Forward Network}
\acrodef{udp}[UDP]{User Datagram Protocol}
\acrodef{fifo}[FIFO]{First-In-First-Out}
\acrodef{gmr}[GMR]{General Motion Retargeting}
\acrodef{rfi}[RFI]{Random Force Injection}
\acrodef{com}[CoM]{Center of Mass}

\crefname{algorithm}{Alg.}{Algs.}
\Crefname{algocf}{Algorithm}{Algorithms}
\crefname{section}{Sec.}{Secs.}
\Crefname{section}{Section}{Sections}
\crefname{table}{Tab.}{Tabs.}
\Crefname{table}{Table}{Tables}
\crefname{figure}{Fig.}{Fig.}
\Crefname{figure}{Figure}{Figure}

\newcolumntype{Y}{>{\centering\arraybackslash}X}
\newcolumntype{L}{>{\raggedright\arraybackslash}X}

\begin{document}
\title{\model: Engineering a Robust, All-Rounder Whole-Body Humanoid Teleoperation System}

\author{
    Yixuan Li\textsuperscript{*,1,2,7} \quad
    Le Ma\textsuperscript{*,2,7} \quad
    Yutang Lin\textsuperscript{*,3,2,6,7,8,10} \quad
    Yushi Du\textsuperscript{4,2,7} \quad
    Mengya Liu\textsuperscript{2,7} \quad
    Kaizhe Hu\textsuperscript{5}
    \vspace{3pt}\\
    Jieming Cui\textsuperscript{3,2,6,7,8,10} \quad
    Yixin Zhu\textsuperscript{\,\textrm{\Letter}\,,6,3,7,8,10} \quad
    Wei Liang\textsuperscript{\,\textrm{\Letter}\,,1,9} \quad
    Baoxiong Jia\textsuperscript{\,\textrm{\Letter}\,,2,7} \quad
    Siyuan Huang\textsuperscript{\,\textrm{\Letter}\,,2,7}
    \vspace{3pt}\\
    \fontsize{8}{8}\selectfont $^{*}$~Equal contribution \quad 
    $\textrm{\Letter}\,$~Corresponding author \quad
    Project Website: \href{https://omniclone.github.io/}{https://omniclone.github.io/}\\
    \fontsize{8}{8}\selectfont \textsuperscript{1} School of Computer Science and Technology, Beijing Institute of Technology \quad
    \fontsize{8}{8}\selectfont \textsuperscript{2} Beijing Institute for General Artificial Intelligence (BIGAI)\\
    \fontsize{8}{8}\selectfont \textsuperscript{3} Institute for AI, Peking University \quad
    \fontsize{8}{8}\selectfont \textsuperscript{4} Department of Electrical and Electronic Engineering, The University of Hong Kong \\
    \fontsize{8}{8}\selectfont \textsuperscript{5} Institute for Interdisciplinary Information Sciences (IIIS), Tsinghua University \quad
    \fontsize{8}{8}\selectfont \textsuperscript{6} School of Psychological and Cognitive Sciences, Peking University \\
    \fontsize{8}{8}\selectfont \textsuperscript{7} State Key Lab of General AI \quad
    \fontsize{8}{8}\selectfont \textsuperscript{8} Beijing Key Laboratory of Behavior and Mental Health, Peking University \\
    \fontsize{8}{8}\selectfont \textsuperscript{9} Yangtze Delta Region Academy of Beijing Institute of Technology \quad
    \fontsize{8}{8}\selectfont \textsuperscript{10} Embodied Intelligence Lab, PKU-Wuhan Institute for Artificial Intelligence
    \vspace{-0.5em}%
}

\twocolumn[{
    \renewcommand\twocolumn[1][]{#1}
    \maketitle
    \centering
    \captionsetup{type=figure}
    \includegraphics[width=\linewidth]{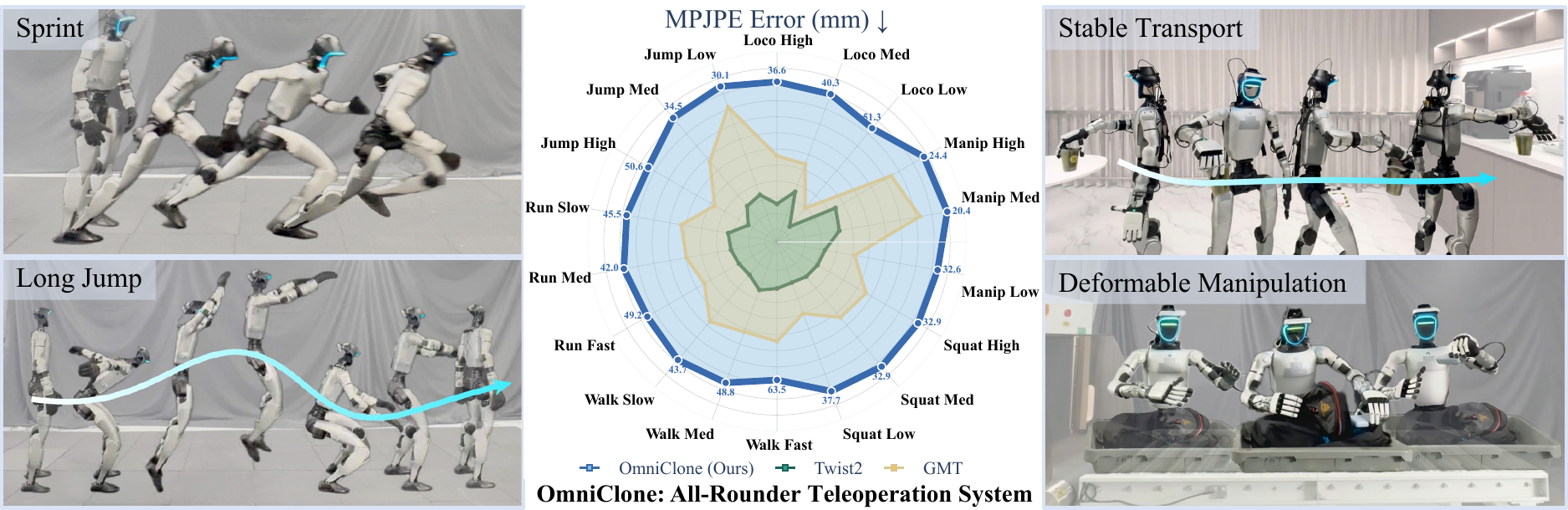}
    \captionof{figure}{\textbf{\model achieves well-balanced, high-fidelity whole-body tracking across all \acs{mpjpe} dimensions on \bench while enabling versatile real-world deployment.} Center: the radar map compares \acs{mpjpe} (mm) across 18 stratified evaluation categories, showing that \model consistently outperforms \acs{sota} baselines GMT and Twist2. Surrounding panels illustrate real-world teleoperation with a unified policy, spanning dynamic whole-body motions (sprint, long jump) and stable, long-horizon dexterous loco-manipulation (transport, deformable object handling).}\vspace{3pt}
    \label{fig:teaser}
}]

\begin{abstract}
Whole-body humanoid teleoperation enables humans to remotely control humanoid robots, serving as both a real-time operational tool and a scalable engine for collecting demonstrations for autonomous learning.
Despite recent advances, existing systems are validated using aggregate metrics that conflate distinct motion regimes, masking critical failure modes. This lack of diagnostic granularity, compounded by tightly coupled and labor-intensive system configurations, hinders robust real-world deployment.
A key open challenge is building a teleoperation system that is simultaneously robust, versatile, and affordable for practical use.
Here we present \model, a whole-body humanoid teleoperation system that achieves high-fidelity, multi-skill control on a single consumer GPU with modest data requirements.
Central to our approach is \bench, a diagnostic benchmark that evaluates policies across stratified motion categories and difficulty levels on unseen motions, exposing the narrow specialization of prior systems. Guided by these diagnostics, we identify an optimized training data recipe and integrate system-level improvements---subject-agnostic retargeting and robust communication---that collectively reduce \ac{mpjpe} by over 66\% while requiring orders-of-magnitude fewer computational resources than comparable methods.
Crucially, \model is control-source-agnostic: a single unified policy supports real-time teleoperation, generated motion playback, and \ac{vla} models, while generalizing across operators of vastly different body proportions.
By uniting diagnostic evaluation with practical engineering, \model provides an accessible foundation for scalable humanoid teleoperation and autonomous learning.
\end{abstract}

\section{Introduction}

Humanoid robots are emerging as versatile, general-purpose platforms capable of executing complex tasks either autonomously~\cite{du2025learning,he2025viral,hu2025robot,jiang2025wholebodyvla,jiang2025uniact,liao2025beyondmimic,lin2026lessmimic,xue2025opening,wang2026humanx} or via teleoperation~\cite{darvish2023teleoperation,he2024learning}. Whole-body teleoperation~\cite{ben2025homie,he2024omnih2o,ji2024exbody2,li2025clone,ze2025twist}, in particular, serves as a foundational technology---enabling both real-time remote intervention and the collection of high-quality demonstrations for scalable imitation learning. Yet this broad operational scope demands the seamless integration of often conflicting capabilities: robustness, stability, dynamic agility, and precise control.

While reliable teleoperation frameworks exist for fixed-base manipulators~\cite{qin2023anyteleop,si2024tilde} and inherently stable quadrupeds~\cite{fu2023deep,ha2025prediction}, achieving comparable maturity for whole-body humanoid teleoperation remains a formidable challenge. Recent methods~\cite{ben2025homie,chen2025gmt,li2025amo,li2025clone,luo2025sonic,ze2025twist2} have markedly expanded the capabilities of humanoid teleoperation, yet these systems still fall short of the robustness and task-agnostic versatility required for real-world deployment. We identify two root causes.

On the \emph{evaluation} side, current methods tend to showcase isolated, high-complexity skills while reporting only coarse aggregate metrics that conflate distinct motion regimes. This masks critical nuances: a policy may maintain low tracking error during a standard upright pose yet exhibit significant instability during a deep squat. By averaging out such failure modes, existing evaluations obscure the narrow specialization of current models and the key bottlenecks that must be resolved for practical, generalizable deployment.

On the \emph{system} side, the hardware and software configurations of current teleoperation systems are highly heterogeneous and tightly coupled with specific methods, hindering real-world reproducibility. Taking \ac{mocap} as an example, platforms like PICO VR headsets~\cite{pico4_ultra} and VICON systems~\cite{vicon_motion_systems} employ proprietary human-scale estimation algorithms that are often opaque to the user. The seemingly minor discrepancies in estimated scale lead to large performance gaps in practice. Compounded by factors such as control frequency and network latency, these systems necessitate labor-intensive calibration for every new operator and each distinct \ac{mocap} setting, severely impeding scalable deployment.

In this work, we adopt a system-level perspective to address these challenges, providing an effective and affordable solution. Rather than proposing yet another task or model in isolation, we begin by constructing \bench, the first comprehensive diagnostic evaluation benchmark for humanoid full-body teleoperation. \bench systematically assesses policy performance under varied workspaces and skill demands---from high-dynamic agile motions to stable, precise quasi-static manipulation---across multiple stratified difficulty levels. As shown in \cref{fig:teaser}, this fine-grained evaluation uncovers persistent skill imbalances in existing systems, arising from both training data imbalance and model design. These limitations are further exacerbated in real-world deployment, where human-scale variabilities, retargeting errors, and data-stream disturbances caused by network fluctuation and latency jointly magnify performance gaps.

Guided by these diagnostics, we present \model, a robust and affordable system for high-fidelity whole-body humanoid teleoperation across a comprehensive range of practical scenarios. \model addresses the narrow coverage of existing models through a high-capacity transformer-based full-body tracking policy trained on a meticulously designed data recipe that balances diverse skills, from high agility to stable manipulation. Beyond policy learning, \model incorporates system-level mechanisms to handle real-world disturbances, including subject-agnostic refined retargeting and robust data communication that mitigate network fluctuation and latency. Crucially, the entire system remains highly accessible, requiring only 30 hours of motion data and a single consumer GPU---orders of magnitude less than comparable methods. We validate \model extensively on \bench, demonstrating significant improvements over \ac{sota} baselines with detailed ablations justifying each design choice. Moreover, \model generalizes reliably across teleoperators ranging from 1.47\,m to 1.94\,m in height, and serves as a control-source-agnostic platform compatible with real-time teleoperation, generated motion playback, and downstream \ac{vla} models. To demonstrate the latter, we train a \ac{vla} policy on \model-collected data, achieving success rates of $85.71\%$ and $80.00\%$ on the `Pick-and-Place' and `Squat to Pick-and-Place' tasks, respectively.

Our contributions can be summarized as follows:
\begin{itemize}[leftmargin=*,noitemsep,nolistsep,topsep=0pt,partopsep=0pt]
    \item We introduce \bench, the first comprehensive diagnostic benchmark for systematically evaluating humanoid full-body teleoperation across diverse motion categories and stratified difficulty levels, providing actionable insights for policy improvement.
    \item We propose \model, a robust, affordable, and easily reproducible system for high-fidelity real-world humanoid teleoperation, integrating a skill-balanced policy with system-level mechanisms for handling deployment disturbances.
    \item We conduct extensive experiments demonstrating that \model significantly outperforms \ac{sota} baselines across all dimensions on \bench and in real-world testing, while further validating its utility as a data collection engine for learning autonomous \ac{vla} policies.
\end{itemize}

\section{Related Work}

\subsection{Dataset and Evaluation in Humanoid Teleoperation}

While numerous recent works~\cite{jiang2025uniact,li2025amo,li2025clone,ze2025twist2,wang2025physhsi} develop custom datasets for humanoid teleoperation, rigorous analysis of how models perform across distinct functional capabilities remains largely absent. Standard evaluation protocols report overall \ac{sr} and tracking errors that conflate distinct movement patterns---from precise manipulation to dynamic locomotion---into single scalar values, obscuring critical performance discrepancies across dynamic regimes and masking specific failure modes. Consequently, most whole-body teleoperation efforts~\cite{chen2025gmt,li2025clone,ze2025twist2} evaluate performance exclusively on self-curated training distributions, where systems may exhibit superiority in isolated skills while suffering significant degradation in other skills. This practice further fails to capture true generalization, as evaluation and training distributions are identical.

To bridge this gap, we introduce \bench, a diagnostic evaluation suite that disentangles performance across multifaceted humanoid capabilities on unseen motions stratified by category and difficulty, thereby approximating real-world deployment conditions. \bench also provides actionable insights for dataset composition, enabling us to curate an optimized training recipe that maximizes tracking fidelity.

\subsection{Whole-Body Humanoid Teleoperation}

Whole-body teleoperation~\cite{fu2024humanplus,he2024learning} has established itself as a cornerstone for collecting dexterous demonstrations. Existing approaches can be broadly categorized into \textit{decoupled} frameworks~\cite{ben2025homie}, which separate upper- and lower-body control for stability, and \textit{unified} frameworks that learn a single full-body policy. Recent research increasingly favors the latter, as coherent whole-body coordination is essential for natural and precise loco-manipulation.

However, the underactuated, floating-base dynamics of humanoid robots make high-fidelity unified control a multifaceted challenge spanning data curation~\cite{ji2024exbody2,luo2025sonic,ze2025twist}, policy architecture~\cite{he2025hover,ji2024exbody2,lu2025mobile,zhang2025unleashing}, and system-level infrastructure~\cite{ze2025twist2}. Despite extensive progress in each domain individually, a critical gap persists: existing systems remain tightly coupled to specific configurations and lack the robustness required for reliable real-world deployment across diverse operators and scenarios.

In this work, we address this gap from a system-engineering perspective, combining diagnostic benchmarking, an optimized data recipe, and a refined deployment infrastructure into a unified, affordable, and robust teleoperation framework.

\section{The \bench Benchmark}\label{sec:bench}

We introduce \bench, a diagnostic evaluation suite designed to quantify the multifaceted capabilities of whole-body teleoperation policies. Below, we describe the benchmark construction and present baseline evaluations that expose the limitations of current systems.

\subsection{Benchmark Construction}

To ensure fair and rigorous evaluation on motions unseen during training, we leverage HY-Motion-$1.0$~\cite{hymotion2025} to generate a comprehensive set of reference motions, which are then retargeted to the humanoid morphology via General Motion Retargeting~(GMR)~\cite{araujo2025retargeting}. The resulting evaluation set spans the full spectrum of humanoid capabilities, from basic locomotion to complex daily manipulation.

Specifically, we construct motions across six functional categories: `Loco-manipulation,' `Manipulation,' `Squatting,' `Walking,' `Running,' and `Jumping,' collectively covering the primary operational domains of teleoperated humanoid robots. Each category comprises $60$ episodes with an average duration of $90$ frames sampled at $30$~Hz. To enable granular diagnosis, we further stratify these motions by \textit{complexity} (High, Medium, Low) and \textit{dynamic intensity} (Fast, Medium, Slow), yielding a fine-grained grid over both skill type and difficulty level. Further details on the evaluation dataset construction are provided in \cref{sec:app:dataset}.

\begin{figure}[t!]
    \centering
    \small
    \includegraphics[width=0.95\linewidth]{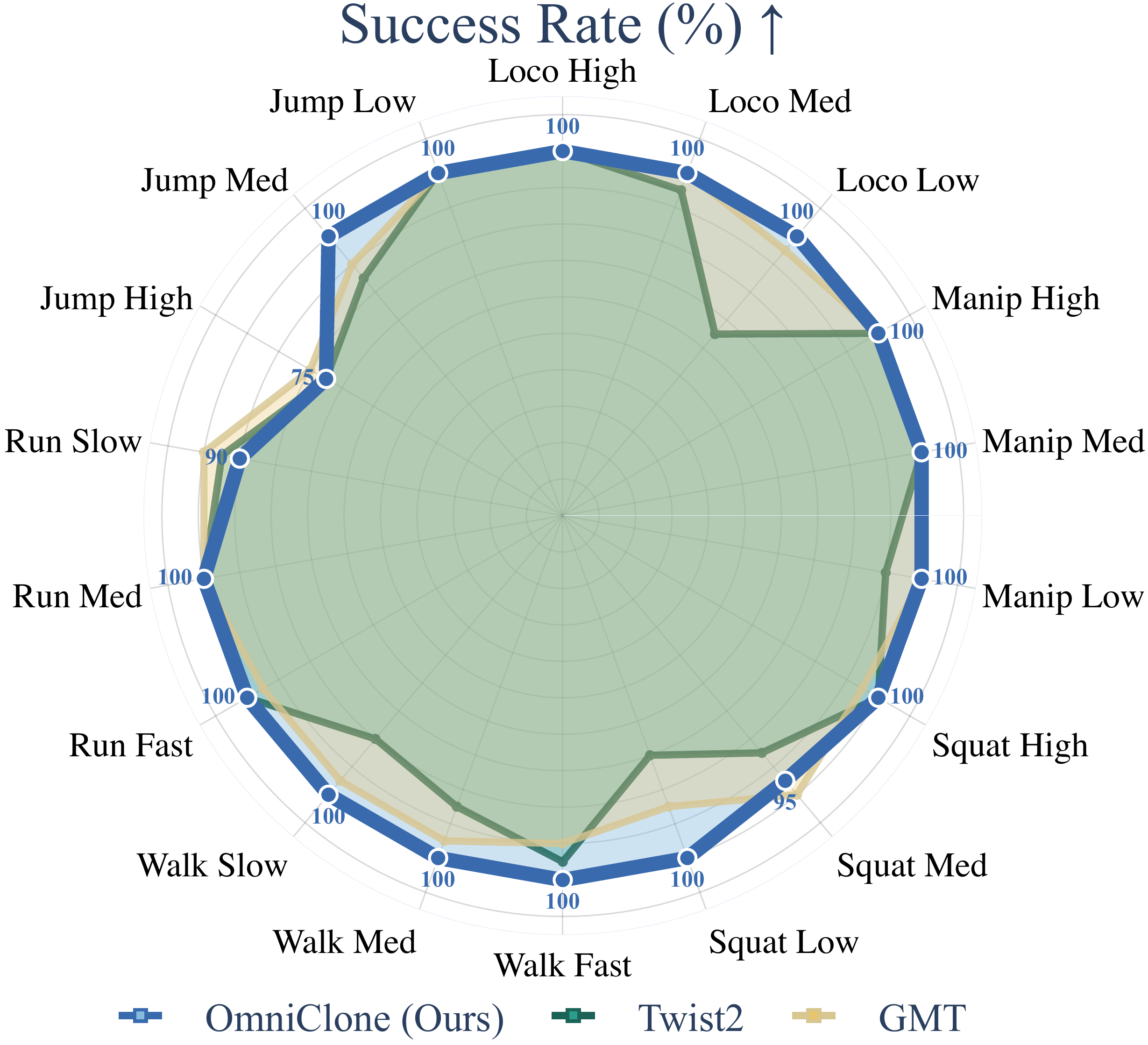}
    \caption{\textbf{\acs{sr} comparison on \bench reveals narrow skill specialization of existing methods.} While GMT and Twist2 suffer notable drops in lower-body agility tasks (deep squatting, low-altitude loco-manipulation) and high-dynamic maneuvers (jumping), \model maintains near-perfect \acs{sr} across all 18 stratified categories. Complementing the \acs{mpjpe} radar map in \cref{fig:teaser}, these results confirm that prior systems' aggregate scores mask significant per-category failures.}
    \label{fig:bench_results}
\end{figure}

\subsection{Baseline Evaluation}

We evaluate representative \ac{sota} methods on \bench, with results shown in \cref{fig:teaser,fig:bench_results}. The evaluation reveals a consistent pattern: existing whole-body teleoperation models remain narrowly specialized. While most systems achieve reasonable performance on manipulation tasks performed from a standard standing stance at medium-to-high heights, they exhibit significant degradation in other regimes---particularly tasks demanding lower-body agility, such as deep squatting or low-altitude loco-manipulation. These findings confirm that current aggregate evaluations mask critical skill imbalances, and provide empirical guidance for the policy and data improvements presented in the following sections.

\section{The \model Teleoperation System}

We present \model, a whole-body humanoid teleoperation system engineered for robust and versatile real-world performance. The framework comprises two integral components: (i) a transformer-based full-body control policy and (ii) an optimized whole-body teleoperation infrastructure. These components work synergistically to enable high-fidelity execution across a diverse spectrum of motions.

\begin{figure*}[t!]
    \centering
    \small
    \includegraphics[width=\linewidth]{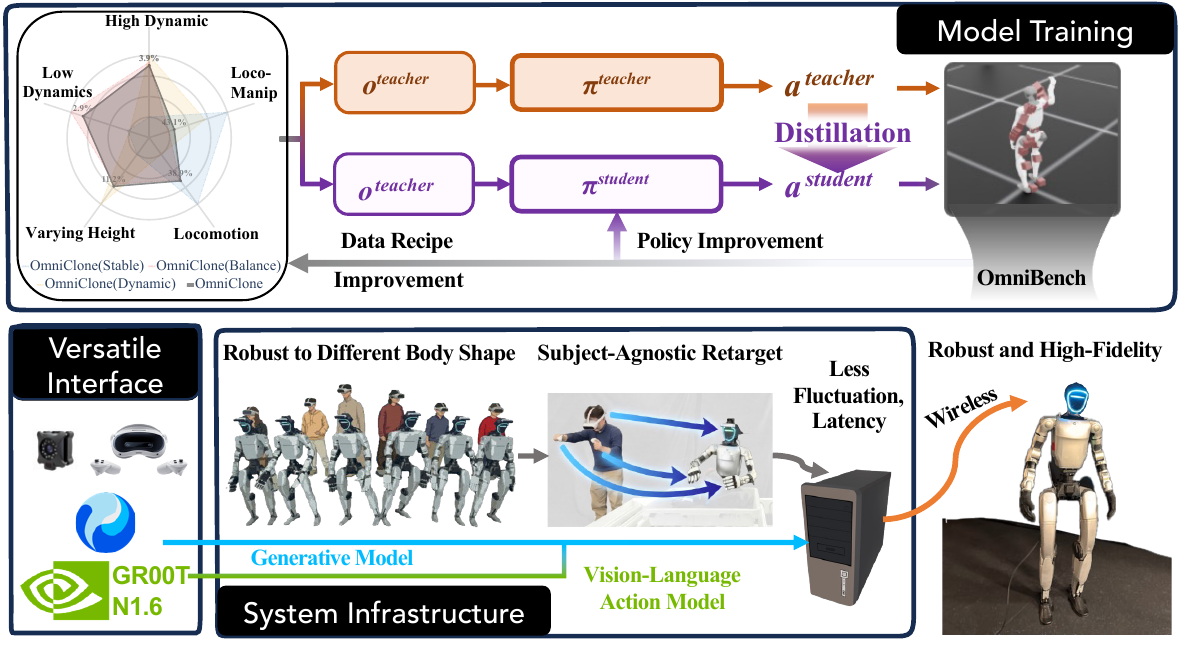}
    \caption{\textbf{Overview of the \model framework, comprising model training (top) and system infrastructure (bottom).} Top: a teacher--student distillation pipeline trained on an optimized data recipe, whose composition is iteratively refined via \bench diagnostics. Bottom: the deployment infrastructure features subject-agnostic retargeting and robust wireless communication, yielding a control-source-agnostic interface compatible with real-time teleoperation, generative motion models, and downstream \acs{vla} planners such as \groot N1.6.}
    \label{fig:pipeline}
\end{figure*}

\subsection{Transformer-Based Whole-Body Tracking Policy}

As established in \cref{sec:bench}, humanoid teleoperation demands versatile task execution characterized by diverse kinematic patterns. Commonly used \ac{mlp} architectures~\cite{he2024omnih2o,zhang2025track} often struggle to capture the complex temporal dependencies inherent in such tasks. We therefore adopt a high-capacity Transformer architecture, whose ability to model long-range sequences is well-suited for processing histories of proprioceptive observations and actions. Its scalability further benefits learning from our large, high-variability training environments.

Building on the demonstrated success of Transformers in humanoid locomotion~\cite{he2025attention,radosavovic2024real}, we implement both the teacher and student policies $\pi$ using a Transformer backbone:
\begin{equation}
    \pi: \mathbb{O} \to \mathbb{A}, \quad \mathbb{A} = \{ \mathbf{a} \mid \mathbf{a} \in \mathbb{R}^n \},
\end{equation}
where $\mathbb{O}$ represents the observation space and $\mathbb{A}$ defines the action space in joint-space coordinates.

We train this policy using the data recipe distilled from our ablation study in \cref{sec:exp_motion_tracking}: $60\%$ manipulation motions, with the remaining $40\%$ balanced between dynamic maneuvers and stable locomotion.

\paragraph*{Teacher policy}
The teacher policy has access to a comprehensive set of joint state observations, including full body orientations and angular velocities, to facilitate high-fidelity behavior learning. The teacher observation vector $\mathbb{O}^{\text{teacher}}$ is defined as:
\begin{equation}
    \mathbb{O}^{\text{teacher}} = [\mathbf{q}, \mathbf{p}, \mathbf{g}, \boldsymbol{\omega}^{\text{root}}, \mathbf{a}^{\text{last}}, \mathbf{q}^{\text{ref}}, \mathbf{p}^{\text{ref}}],
\end{equation}
where $\mathbf{q}$ and $\mathbf{p}$ represent the robot's full kinematic state, comprising joint states (positions and velocities) and body link states (positions, orientations, linear and angular velocities); $\mathbf{g}$ denotes the gravity vector; $\boldsymbol{\omega}^{\text{root}}$ is the angular velocity of the robot's base (pelvis); and $\mathbf{a}^{\text{last}}$ is the previous action. $\mathbf{q}^{\text{ref}}$ and $\mathbf{p}^{\text{ref}}$ correspond to the reference motion's joint states and body link kinematics. All observations are expressed in the robot's local base frame to ensure spatial invariance.

\paragraph*{Student policy}
For robust real-world deployment, we exclude observations that are difficult to estimate reliably on physical hardware, such as the angular velocities of individual body links, and provide $f$ frames of future reference motion. The student observation vector $\mathbb{O}^{\text{student}}$ is:
\begin{equation}
    \mathbb{O}^{\text{student}} = [\mathbf{q}, \mathbf{g}, \boldsymbol{\omega}^{\text{root}}, \mathbf{a}^{\text{last}}, \mathbf{p}^{\text{ref}}_{t:t+f}],
\end{equation}
where $\mathbf{p}^{\text{ref}}$ denotes the reference body state command, including the positions and orientations of key body links and the linear velocity of the robot base.

\paragraph*{Network architecture and reward formulation}
Details of the network architecture and reward formulation are provided in \cref{sec:app:model_structure}.

\subsection{Optimized Whole-Body Teleoperation Infrastructure}

Beyond policy learning, robust real-world performance hinges on the deployment infrastructure itself. We identify two critical system-level factors that degrade teleoperation fidelity and address each with targeted mechanisms.

\paragraph*{Retargeting discrepancies}
Morphological inconsistencies introduce significant noise into the retargeting pipeline. These discrepancies arise from two sources: systematic variations across different \ac{mocap} hardware configurations and anthropometric variability among distinct human subjects using the same capture system (\eg, PICO~\cite{pico4_ultra}). We empirically quantify these discrepancies, observing maximum deviations of approximately $20$~cm, which translate to an \ac{mpjpe} increase of $\approx 20$~mm when reproduced in simulation.

To address this, we propose a subject-agnostic refined retargeting pipeline. We compute a dynamic scaling factor from the initial calibration frame and the humanoid's kinematics, then use it to rescale the raw \ac{mocap} data to match the humanoid's morphology. This eliminates the need for per-operator manual calibration while effectively mitigating geometric discrepancies in the retargeting process.

\begin{table*}[t!]
    \centering
    \small
    \setlength{\tabcolsep}{3pt}
    \caption{\textbf{Quantitative comparison with baselines and ablation on training data recipes, evaluated on \bench.} All results are stratified by motion category and difficulty level, reporting \acs{sr} (\%) and \acs{mpjpe} (mm). \model achieves the best overall balance across all categories, while the ablation variants illustrate how data composition steers policy behavior. Best \acs{mpjpe} per category is in bold.}
    \label{tab:combined_full}
    \begin{subtable}{\linewidth}
        \centering
        \caption{Locomotion and manipulation tasks, stratified by workspace height (High, Medium, Low).}
        \label{tab:sub_loco_manip}
        \resizebox{\linewidth}{!}{
            \begin{tabular}{l @{\hspace{1em}} cc cc cc @{\hspace{1em}} cc cc cc @{\hspace{1em}} cc cc cc}
                \toprule
                \multirow{3}{*}{\textbf{Method}} 
                & \multicolumn{6}{c}{\textbf{Loco-Manip}} 
                & \multicolumn{6}{c}{\textbf{Manip}} 
                & \multicolumn{6}{c}{\textbf{Squat}} \\ 
                \cmidrule(r{1em}){2-7} \cmidrule(r{1em}){8-13} \cmidrule{14-19}
                & \multicolumn{2}{c}{High} & \multicolumn{2}{c}{Medium} & \multicolumn{2}{c}{Low}
                & \multicolumn{2}{c}{High} & \multicolumn{2}{c}{Medium} & \multicolumn{2}{c}{Low}
                & \multicolumn{2}{c}{High} & \multicolumn{2}{c}{Medium} & \multicolumn{2}{c}{Low} \\
                \cmidrule(r){2-3} \cmidrule(lr){4-5} \cmidrule(lr){6-7}
                \cmidrule(r){8-9} \cmidrule(lr){10-11} \cmidrule(lr){12-13}
                \cmidrule(r){14-15} \cmidrule(lr){16-17} \cmidrule(l){18-19}
                & SR $\uparrow$ & MPJPE $\downarrow$ & SR $\uparrow$ & MPJPE $\downarrow$ & SR $\uparrow$ & MPJPE $\downarrow$
                & SR $\uparrow$ & MPJPE $\downarrow$ & SR $\uparrow$ & MPJPE $\downarrow$ & SR $\uparrow$ & MPJPE $\downarrow$
                & SR $\uparrow$ & MPJPE $\downarrow$ & SR $\uparrow$ & MPJPE $\downarrow$ & SR $\uparrow$ & MPJPE $\downarrow$ \\
                \midrule
                GMT~\cite{chen2025gmt} & $100$ & $128.4$ & $100$ & $132.5$ & $95$ & $180.5$ & $100$ & $71.4$ & $100$ & $54.7$ & $100$ & $137.0$ & $95$ & $107.4$ & $100$ & $112.4$ & $85$ & $140.1$\\
                Twist$2$~\cite{ze2025twist2} & $100$ & $188.0$ & $95$ & $168.9$ & $65$ & $210.5$ & $100$ & $151.1$ & $100$ & $156.3$ & $90$ & $174.6$ & $100$ & $176.9$ & $85$ & $178.5$ & $70$ & $181.1$\\
                \midrule
                \model\textsuperscript{MLP} & $85$ & $58.90$ & $65$ & $75.4$ & $65$ & $73.4$ & $100$ & $36.3$ & $100$ & $37.1$ & $100$ & $47.9$ & $100$ & $49.5$ & $100$ & $47.3$ & $95$ & $54.7$ \\
                \model & $100$ & $36.6$ & $100$ & $40.3$ & $100$ & $51.3$ & $100$ & $24.4$ & $100$ & $20.4$ & $100$ & $\textbf{32.6}$ & $100$ & $32.9$ & $95$ & $32.9$ & $100$ & $37.7$ \\
                \midrule
                \model\textsuperscript{Balance} & $100$ & $\textbf{35.3}$ & $100$ & $\textbf{39.9}$ & $100$ & $48.1$ & $100$ & $\textbf{19.0}$ & $100$ & $\textbf{18.4}$ & $100$ & $33.3$ & $100$ & $\textbf{32.0}$ & $100$ & $\textbf{31.4}$ & $90$ & $\textbf{37.1}$ \\
                \model\textsuperscript{Stable} & $100$ & $37.2$ & $95$ & $43.7$ & $100$ & $50.2$ & $100$ & $22.2$ & $100$ & $20.8$ & $100$ & $38.4$ & $100$ & $39.7$ & $100$ & $38.1$ & $95$ & $45.2$ \\
                \model\textsuperscript{Dynamic} & $30$ & $78.1$ & $20$ & $96.8$ & $10$ & $92.8$ & $75$ & $53.9$ & $75$ & $53.7$ & $25$ & $75.4$ & $35$ & $72.9$ & $45$ & $65$ & $35$ & $71.6$ \\
                \bottomrule
            \end{tabular}%
        }%
    \end{subtable}%
    \vspace{3pt}\\%
    \begin{subtable}{\linewidth}
        \centering
        \caption{Agile motion tasks, stratified by dynamic intensity (Fast, Medium, Slow) for Walk and Run, and by height for Jump.}
        \label{tab:sub_agile}
        \resizebox{\linewidth}{!}{
            \begin{tabular}{l @{\hspace{1em}} cc cc cc @{\hspace{1em}} cc cc cc @{\hspace{1em}} cc cc cc}
                \toprule
                \multirow{3}{*}{\textbf{Method}} 
                & \multicolumn{6}{c}{\textbf{Walk}} 
                & \multicolumn{6}{c}{\textbf{Run}} 
                & \multicolumn{6}{c}{\textbf{Jump}} \\ 
                \cmidrule(r{1em}){2-7} \cmidrule(r{1em}){8-13} \cmidrule{14-19}
                & \multicolumn{2}{c}{Fast} & \multicolumn{2}{c}{Medium} & \multicolumn{2}{c}{Slow}
                & \multicolumn{2}{c}{Fast} & \multicolumn{2}{c}{Medium} & \multicolumn{2}{c}{Slow}
                & \multicolumn{2}{c}{High} & \multicolumn{2}{c}{Medium} & \multicolumn{2}{c}{Low} \\
                \cmidrule(r){2-3} \cmidrule(lr){4-5} \cmidrule(lr){6-7}
                \cmidrule(r){8-9} \cmidrule(lr){10-11} \cmidrule(lr){12-13}
                \cmidrule(r){14-15} \cmidrule(lr){16-17} \cmidrule(l){18-19}
                & SR $\uparrow$ & MPJPE $\downarrow$ & SR $\uparrow$ & MPJPE $\downarrow$ & SR $\uparrow$ & MPJPE $\downarrow$
                & SR $\uparrow$ & MPJPE $\downarrow$ & SR $\uparrow$ & MPJPE $\downarrow$ & SR $\uparrow$ & MPJPE $\downarrow$
                & SR $\uparrow$ & MPJPE $\downarrow$ & SR $\uparrow$ & MPJPE $\downarrow$ & SR $\uparrow$ & MPJPE $\downarrow$ \\
                \midrule
                GMT~\cite{chen2025gmt} & $90$ & $111.4$ & $95$ & $116.0$ & $95$ & $105.5$ & $95$ & $130.9$ & $100$ & $120.8$ & $100$ & $114.1$ & $80$ & $145.6$ & $90$ & $105.3$ & $100$ & $57.1$ \\
                Twist$2$~\cite{ze2025twist2} & $95$ & $176.9$ & $85$ & $171.2$ & $80$ & $181.6$ & $100$ & $180.4$ & $100$ & $176.9$ & $95$ & $173.4$ & $75$ & $193.8$ & $85$ & $177.2$ & $100$ & $173.0$ \\
                \midrule
                \model\textsuperscript{MLP} & $20$ & $111.7$ & $35$ & $95.6$ & $85$ & $64.1$ & $90$ & $82.0$ & $80$ & $69.8$ & $70$ & $71.1$ & $75$ & $81.4$ & $85$ & $62.9$ & $75$ & $58.9$ \\
                \model & $100$ & $63.5$ & $100$ & $48.8$ & $100$ & $43.7$ & $100$ & $49.2$ & $100$ & $42.0$ & $90$ & $45.5$ & $75$ & $50.6$ & $100$ & $\textbf{34.5}$ & $100$ & $30.1$ \\
                \midrule
                \model\textsuperscript{Balance} & $100$ & $\textbf{63.0}$ & $95$ & $\textbf{46.4}$ & $95$ & $\textbf{40.8}$ & $90$ & $\textbf{45.7}$ & $100$ & $\textbf{38.3}$ & $100$ & $\textbf{41.6}$ & $90$ & $\textbf{50.0}$ & $100$ & $35.3$ & $100$ & $\textbf{29.8}$\\
                \model\textsuperscript{Stable} & $100$ & $70.3$ & $100$ & $49.3$ & $100$ & $43.8$ & $100$ & $51.4$ & $100$ & $42.4$ & $95$ & $43.6$ & $80$ & $57.1$ & $100$ & $44.5$ & $100$ & $39.6$\\
                \model\textsuperscript{Dynamic} & $90$ & $64.8$ & $100$ & $48.6$ & $100$ & $44.1$ & $100$ & $47.7$ & $100$ & $38.3$ & $95$ & $41.8$ & $85$ & $48.8$ & $10$ & $74.9$ & $45$ & $63.9$ \\
                \bottomrule
            \end{tabular}%
        }%
    \end{subtable}%
\end{table*}%

\paragraph*{Data fluctuation and latency}
In real-world deployment, control signal fluctuations degrade motion smoothness, while network latency increases the cognitive load on the teleoperator. We address both challenges with a queue-based data manager that buffers $f$ frames of retargeted motion in a FIFO queue at a fixed update frequency. The manager employs a zero-order hold strategy, retaining the previous observation during signal interruptions to ensure smooth execution. We further adopt UDP for communication between the retargeting server and policy server (supporting both remote and onboard configurations) for its superior transmission efficiency. Together, these mechanisms yield a low end-to-end latency of approximately $80$~ms.

\subsection{Implementation Details}

We train our policy in Isaac Lab~\cite{mittal2025isaac} on a single NVIDIA RTX $4090$ GPU. The teacher policy converges in approximately $60$ hours, spanning ${\sim}33$K \ac{ppo} steps with $4{,}096$ parallel environments. The student policy is then distilled from the teacher via DAgger in approximately $22$ hours under the same configuration. In total, the system requires only ${\sim}80$ GPU hours on consumer hardware---orders of magnitude less than comparable methods---underscoring the affordability of \model.

\section{Simulations}

We conduct comprehensive simulations in MuJoCo~\cite{todorov2012mujoco} to evaluate the motion tracking capabilities of \model on the \bench suite. We report both \ac{sr} and \ac{mpjpe} (in mm) across a diverse spectrum of skills at varying difficulty levels. The primary objective is to quantitatively analyze how distinct training data recipes dictate policy performance, thereby distilling actionable insights for policy learning.

\begin{figure}[t!]
    \centering
    \small
    \includegraphics[width=\linewidth]{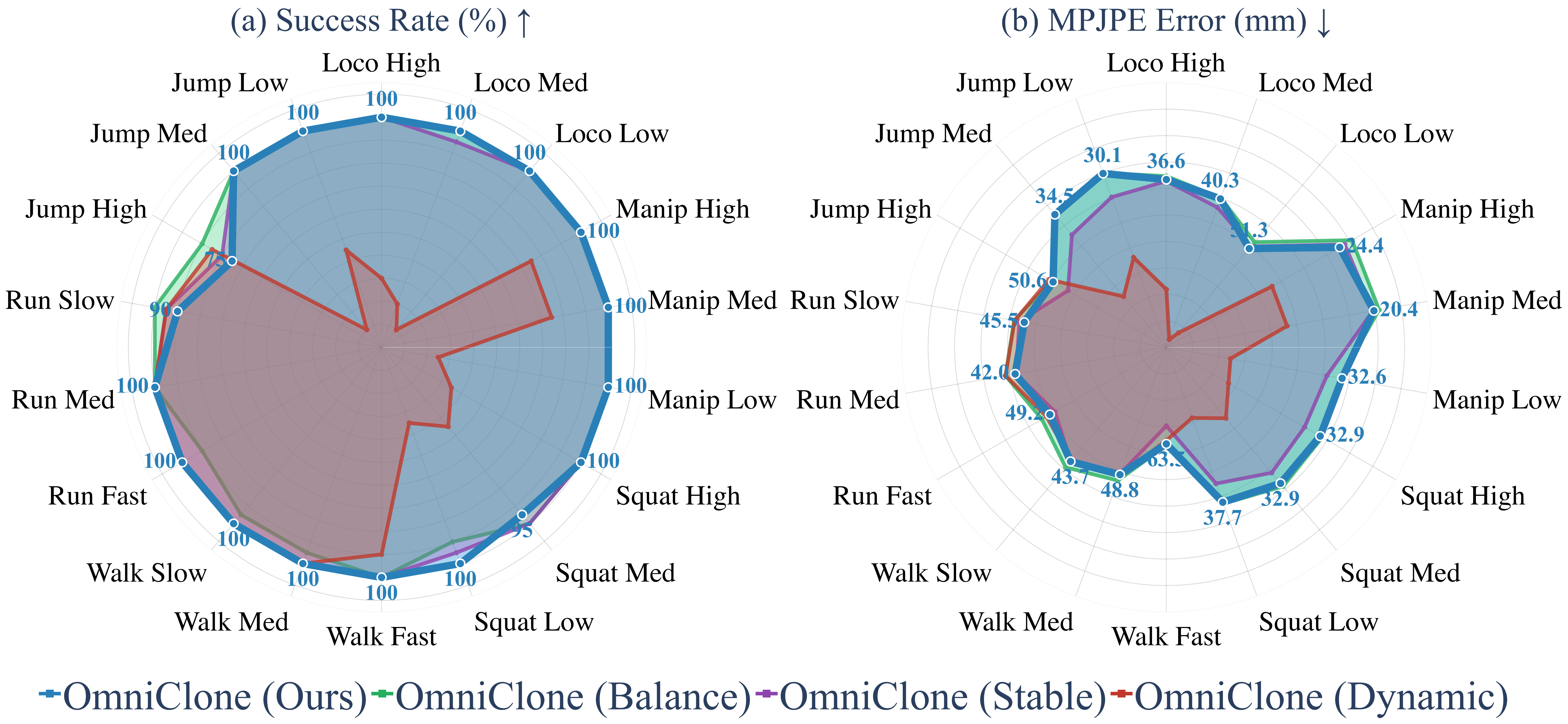}
    \caption{\textbf{Ablations on training data recipes show how data composition steers policy behavior.} \model\textsuperscript{Dynamic} excels at agile motions but collapses on manipulation and squatting; \model\textsuperscript{Stable} recovers stability at the cost of dynamic agility; \model\textsuperscript{Balance} improves breadth across both regimes. The final \model recipe, obtained by filtering biased sequences from \model\textsuperscript{Balance}, achieves the best overall trade-off in both (a) \acs{sr} and (b) \acs{mpjpe}.}
    \label{fig:ablation}
\end{figure}

\begin{figure*}[b!]
    \centering
    \small
    \begin{subfigure}[b]{0.33\linewidth}
        \includegraphics[width=\linewidth]{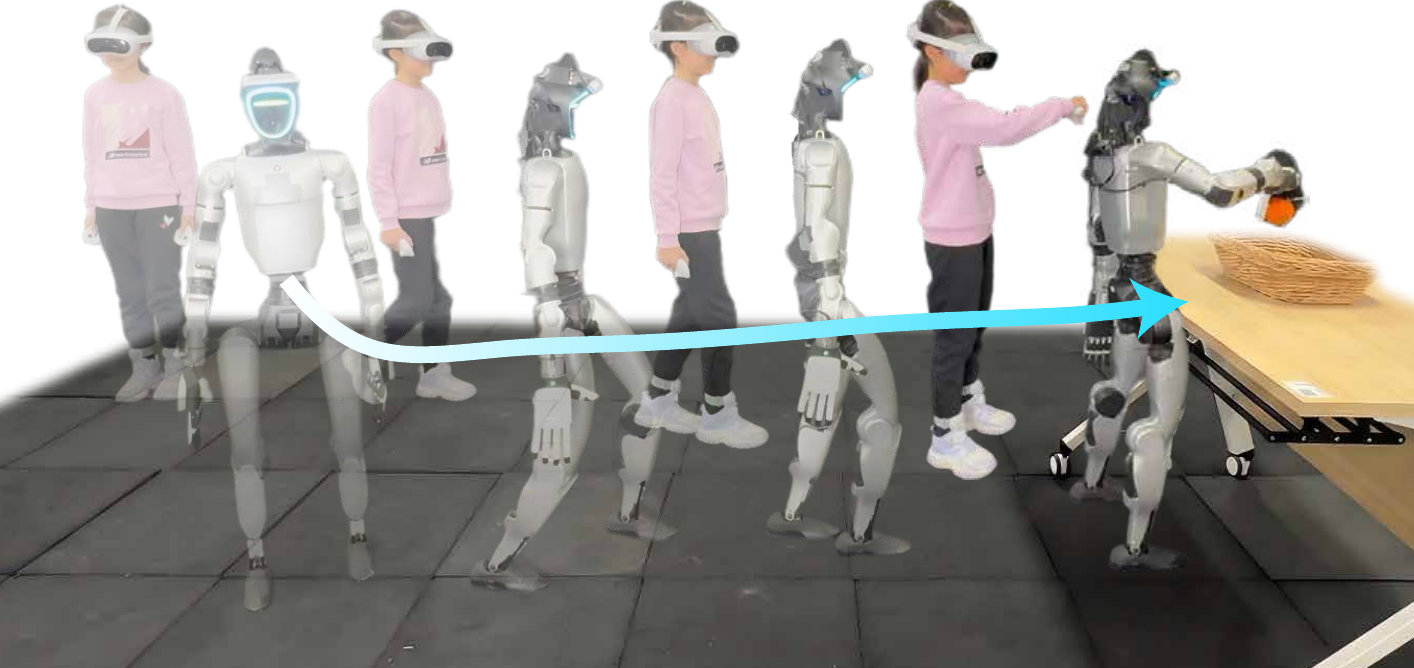}
        \caption{Operator A (height: $147$~cm)}
    \end{subfigure}%
    \begin{subfigure}[b]{0.33\linewidth}
        \includegraphics[width=\linewidth]{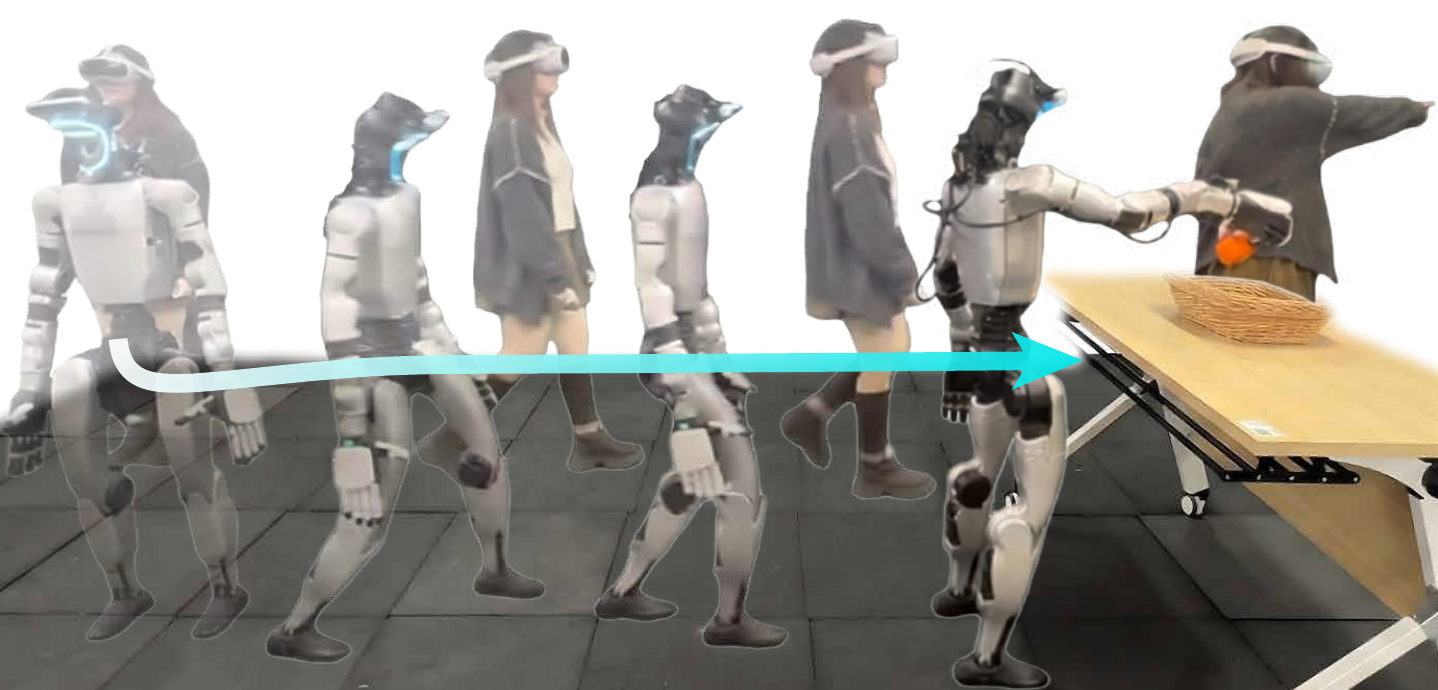}
        \caption{Operator B (height: $154$~cm)}
    \end{subfigure}%
    \begin{subfigure}[b]{0.33\linewidth}
        \includegraphics[width=\linewidth]{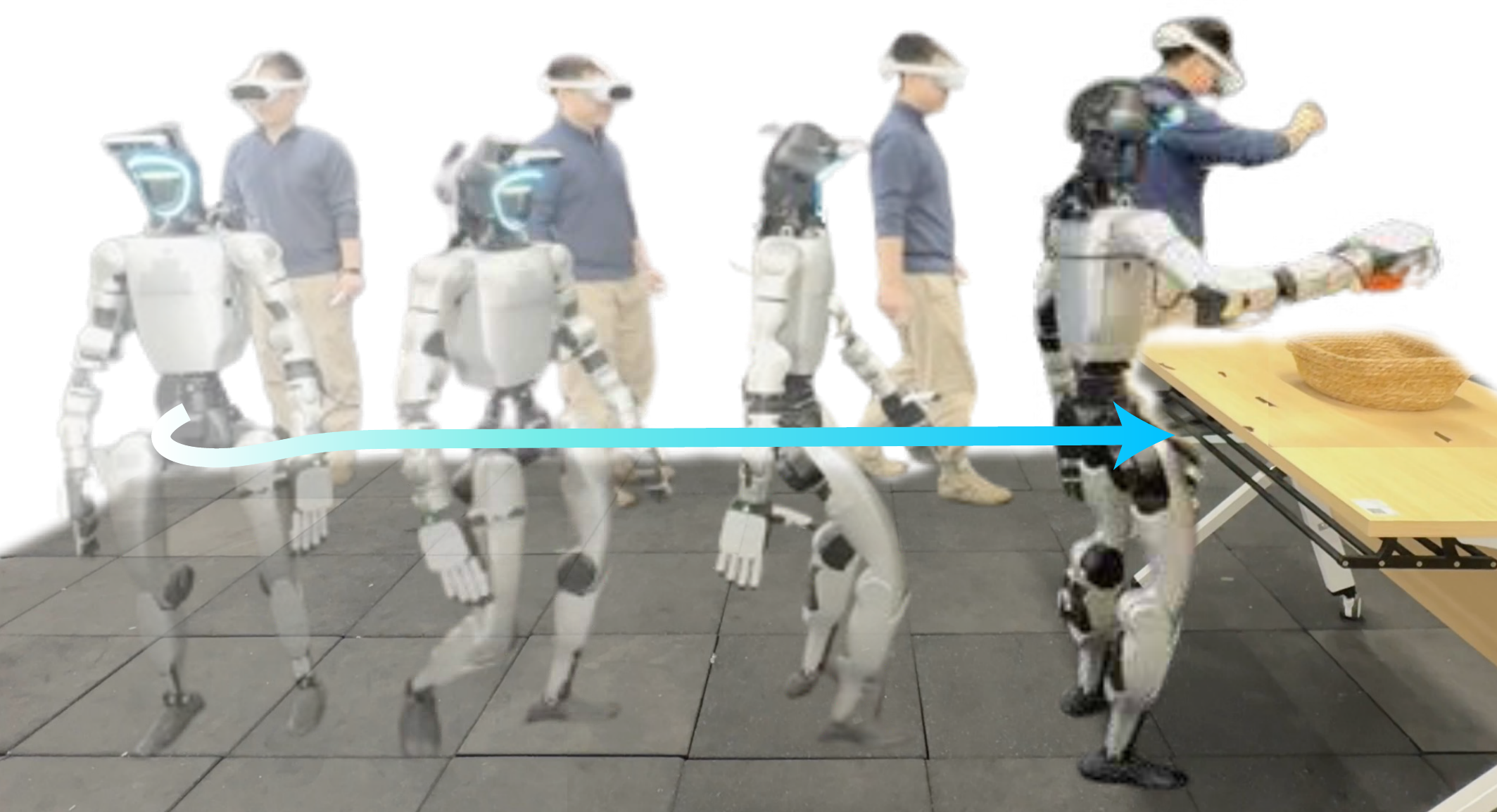}
        \caption{Operator C (height: $165$~cm)}
    \end{subfigure}%
    \\
    \begin{subfigure}[b]{0.33\linewidth}
        \includegraphics[width=\linewidth]{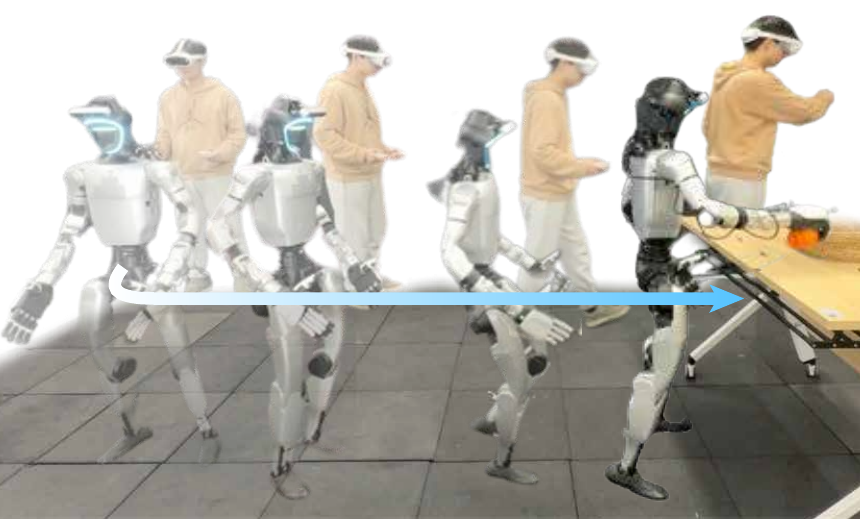}
        \caption{Operator D (height: $175$~cm)}
    \end{subfigure}%
    \begin{subfigure}[b]{0.33\linewidth}
        \includegraphics[width=\linewidth]{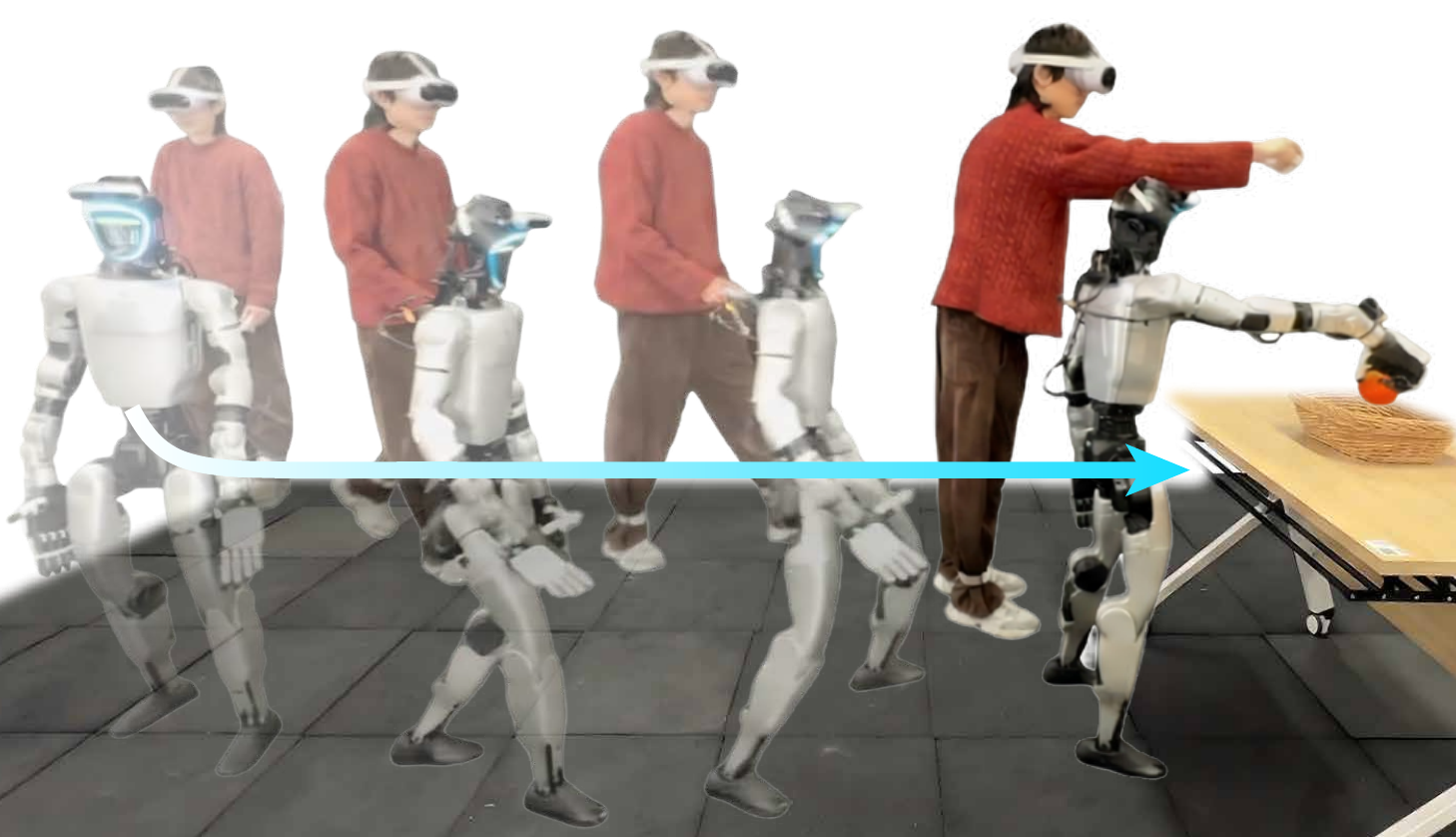}
        \caption{Operator E (height: $182$~cm)}
    \end{subfigure}%
    \begin{subfigure}[b]{0.33\linewidth}
        \includegraphics[width=\linewidth]{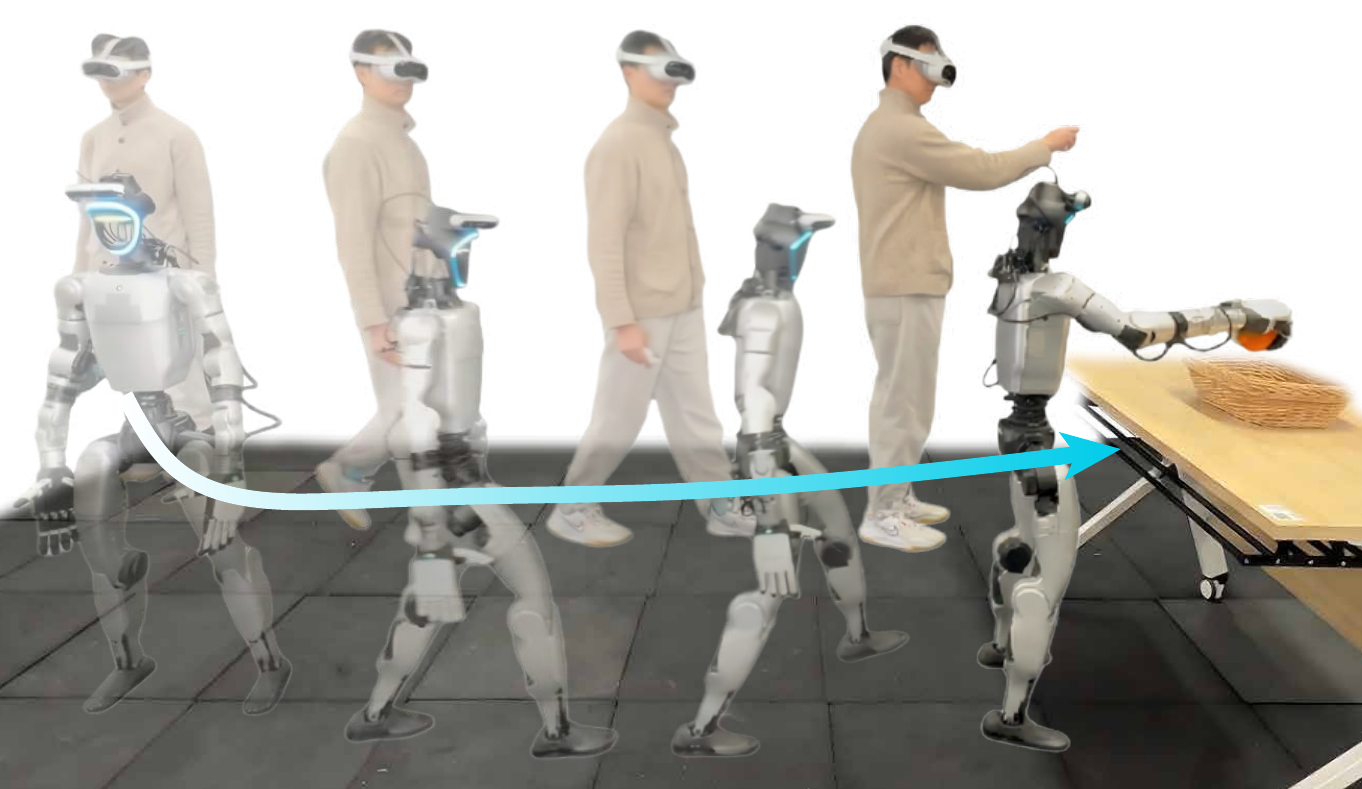}
        \caption{Operator F (height: $194$~cm)}
    \end{subfigure}%
    \caption{\textbf{\model generalizes reliably across teleoperators of vastly different heights.} Six participants ($1.47$\,m--$1.94$\,m) each perform a composite loco-manipulation task (walk, stabilize, pick-and-place). Despite a $47$~cm height span, the system maintains consistent stability and task success throughout, with all novice operators completing the task within $5$--$7$ practice trials.}
    \label{fig:generalize}
\end{figure*}

\subsection{Training Data Recipe}

We curate our training dataset by aggregating motion data from the AMASS~\cite{mahmood2019amass} and LAFAN~\cite{harvey2020robust} datasets. The composition of the final dataset is illustrated in \cref{fig:pipeline}. Since the majority of the data involves manipulation in a standing pose ($\approx 60\%$), the figure specifically highlights the distribution of distinct dynamic behaviors to better visualize the dataset's diversity. A detailed breakdown of the data composition is provided in \cref{sec:app:dataset}.

\subsection{Reference Motion Tracking}\label{sec:exp_motion_tracking}

\paragraph*{Baselines and ablation variants}
We compare \model against two \ac{sota} full-body tracking baselines using their pre-trained checkpoints: GMT~\cite{chen2025gmt} and Twist2~\cite{ze2025twist2}. To validate our data recipe design, we further evaluate several ablation variants whose data compositions are visualized in \cref{fig:ablation}:
\begin{itemize}[leftmargin=*,noitemsep,nolistsep,topsep=0pt,partopsep=0pt]
    \item \model\textsuperscript{Dynamic}: trained exclusively on high-dynamic maneuvers.
    \item \model\textsuperscript{Stable}: augmented with stable motions such as standard locomotion and loco-manipulation.
    \item \model\textsuperscript{Balance}: trained on a comprehensive mixture incorporating additional dynamic motion sources.
    \item \model: our final model, trained on the optimized data mixture from \model\textsuperscript{Balance} after rigorous noise filtering.
    \item \model\textsuperscript{MLP}: an architectural ablation replacing the Transformer backbone with a standard \ac{mlp} policy.
\end{itemize}

\paragraph*{Comparison with baselines}
We evaluate motion tracking performance across diverse simulation tasks, with results presented in \cref{tab:sub_agile,tab:sub_loco_manip,fig:teaser,fig:bench_results}. The quantitative metrics in \cref{tab:sub_agile,tab:sub_loco_manip} demonstrate that \model achieves superior tracking performance across most tasks. The radar maps in \cref{fig:teaser,fig:bench_results} further visualize this comparison in terms of \ac{sr} and \ac{mpjpe}, showing that \model attains a well-balanced performance profile across multi-faceted tasks while substantially improving overall tracking fidelity relative to baselines.

\paragraph*{Ablation on data recipe}
We analyze the impact of training data composition on policy behavior, with the distinct recipes detailed in \cref{fig:pipeline} and corresponding performance comparisons in \cref{fig:ablation}.

\begin{figure*}[t!]
    \centering
    \small
    \begin{subfigure}[b]{0.5\linewidth}
        \includegraphics[width=\linewidth]{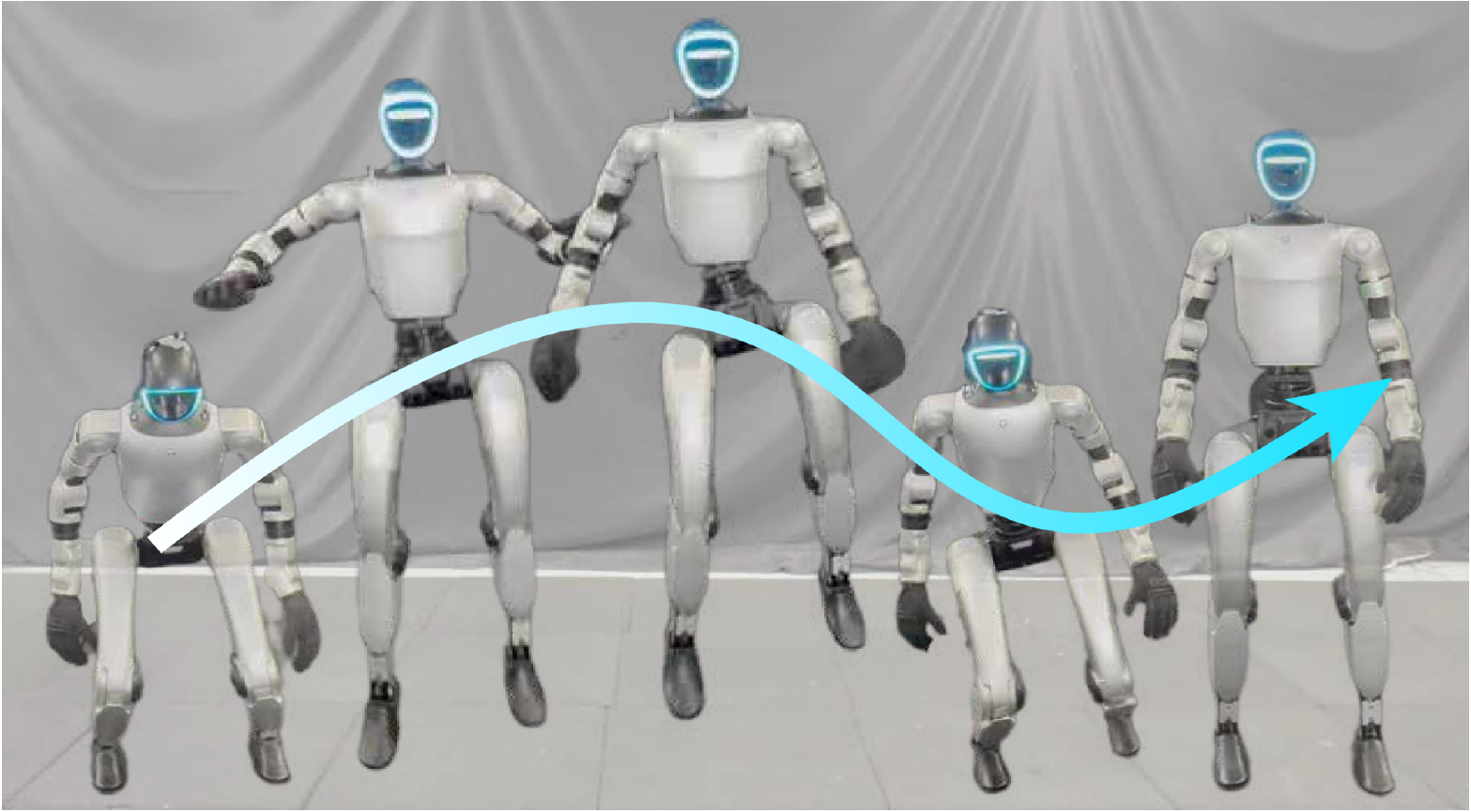}
        \caption{Jump in place}
    \end{subfigure}%
    \begin{subfigure}[b]{0.5\linewidth}
        \includegraphics[width=\linewidth]{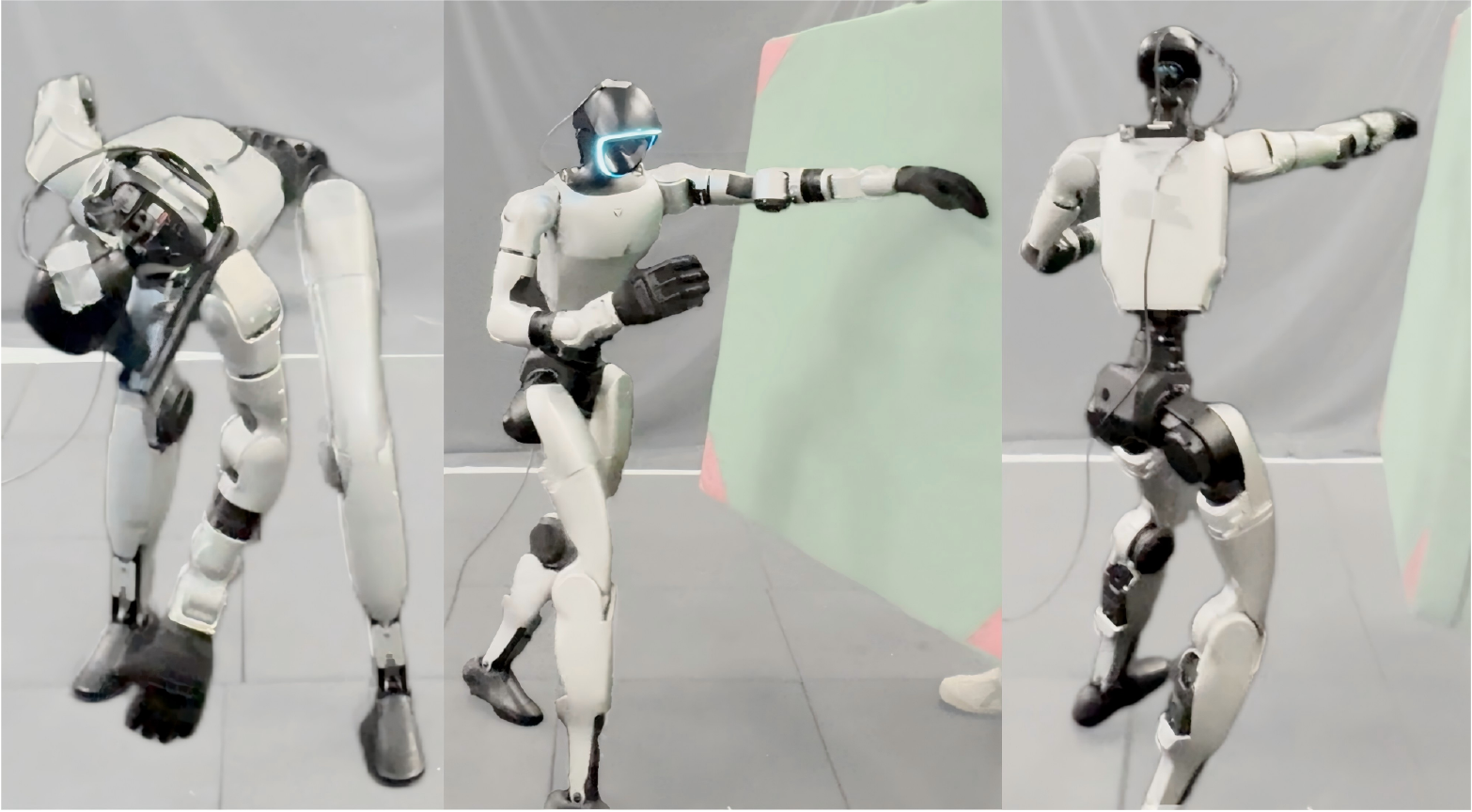}
        \caption{Stretch and punch}
    \end{subfigure}%
    \caption{\textbf{Zero-shot playback of generated motions establishes the performance upper bound of the control policy.} By replaying synthetic motions from HY-Motion-1.0 in the absence of system-induced noise (retargeting artifacts, network latency), we isolate the policy's intrinsic tracking fidelity. Together with the sprint and long-jump results in the left panels of \cref{fig:teaser}, these results confirm that the end-to-end system's residual errors are primarily driven by upstream infrastructure bottlenecks rather than policy capacity.}
    \label{fig:replay}
\end{figure*}

\begin{figure*}[t!]
    \centering
    \small
    \begin{subfigure}[b]{\linewidth}
        \includegraphics[width=\linewidth]{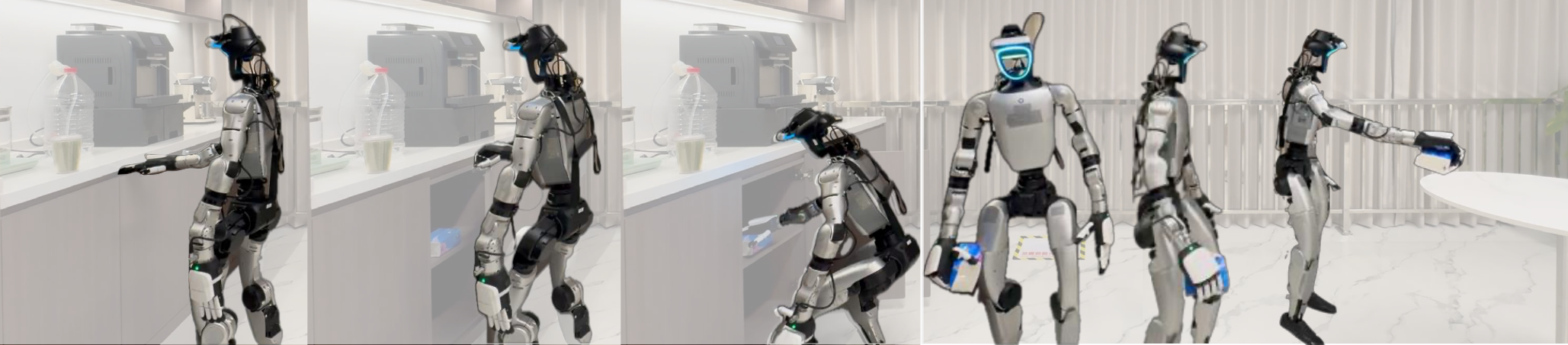}
        \caption{Long-horizon composite task: open cabinet, retrieve tissue, and place on table.}
    \end{subfigure}%
    \\%
    \begin{subfigure}[b]{0.5\linewidth}
        \includegraphics[width=\linewidth]{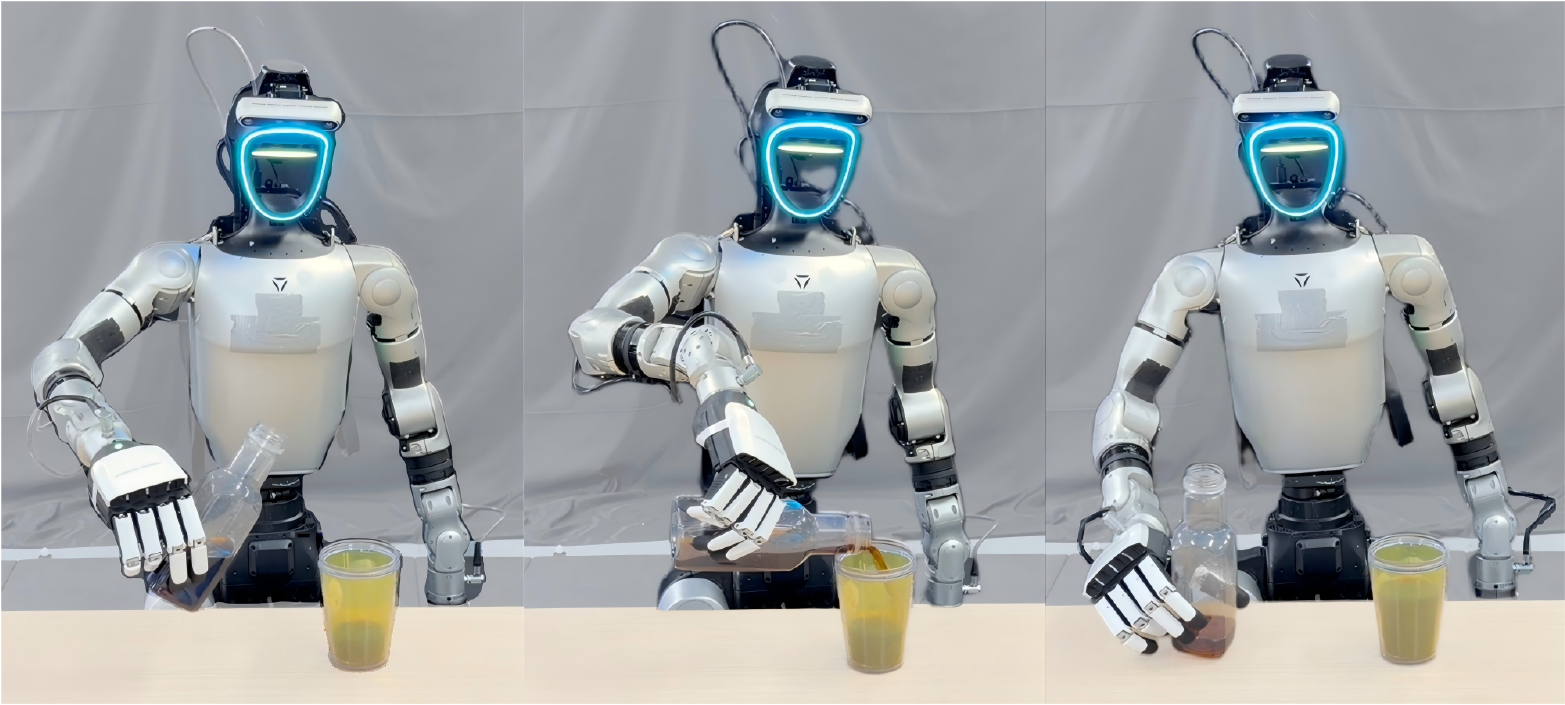}
        \caption{Serving tea without spillage.}
    \end{subfigure}%
    \begin{subfigure}[b]{0.5\linewidth}
        \includegraphics[width=\linewidth]{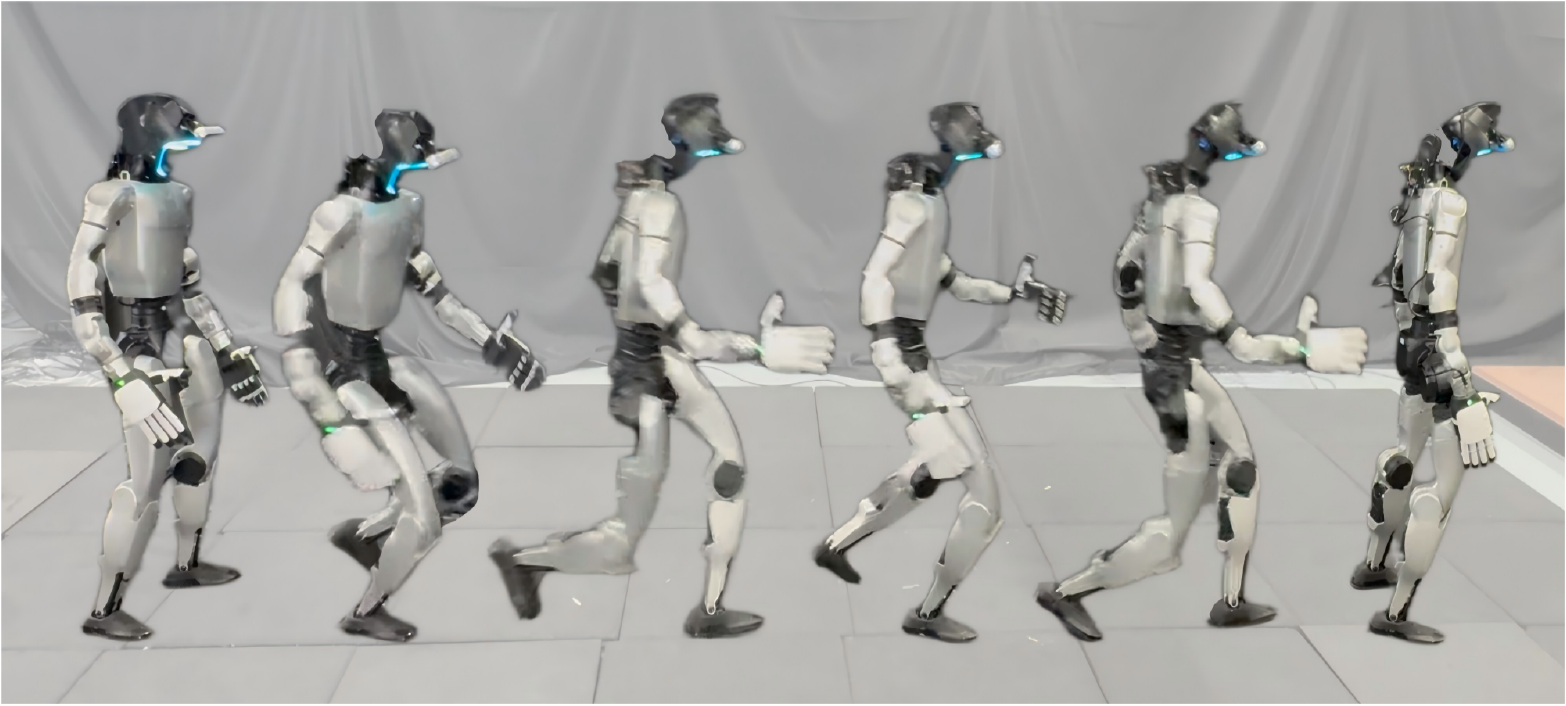}
        \caption{Dynamic running.}
    \end{subfigure}%
    \caption{\textbf{A single unified policy handles the full spectrum of real-world teleoperation, from precise quasi-static manipulation to highly dynamic locomotion.} (a) The humanoid is teleoperated to open a cabinet, retrieve a tissue, and place it on a table, exercising articulated object interaction and coordinated whole-body motion. (b) Stable liquid transport with fine-grained force control is demonstrated by serving tea without spillage. (c) High-speed running further validates the policy's dynamic agility. All tasks are executed via real-time teleoperation without policy switching.}
    \label{fig:tracking}
\end{figure*}

\model\textsuperscript{Dynamic}, trained on a dataset heavily skewed towards high-dynamic motions and varying heights, can execute explosive maneuvers such as jumping but lacks stable locomotion data. This coverage gap prevents the model from learning robust stabilization strategies, resulting in significant instability during precise manipulation tasks. In contrast, \model\textsuperscript{Stable} incorporates a larger proportion of locomotion and loco-manipulation data, yielding improved performance on stable-oriented tasks like squatting and loco-manipulation. By further increasing the proportion of dynamic data, \model\textsuperscript{Balance} achieves more balanced performance across both dynamic and stable regimes, as shown in \cref{fig:ablation}. However, we observed that certain raw dynamic sequences introduce tracking bias during real-world deployment. Filtering these from the \model\textsuperscript{Balance} dataset yields the final, optimized recipe for \model.

A key insight from this analysis is that, although standing manipulation constitutes the majority of the training data, the system's behavioral characteristics are highly sensitive to the composition of the remaining fraction. Subtle adjustments to the proportions of dynamic and locomotion data, combined with filtering of biased sequences, can effectively steer the policy towards distinct behaviors and capabilities.

\paragraph*{Ablation on model architecture}
We quantify the performance differential between the \ac{mlp} and Transformer backbones. As shown in \cref{tab:combined_full}, \model consistently outperforms \model\textsuperscript{MLP} across all metrics, achieving higher \ac{sr} and lower \ac{mpjpe} under identical training settings. We attribute this gain to the Transformer's ability to model long-range temporal dependencies and its enhanced scalability when learning from massive, diverse skill datasets. Comparisons with baseline models further demonstrate that, even when controlling for the model backbone, our data recipe refined using \bench significantly enhances policy learning efficacy.

\section{Experiments}

We conduct real-world experiments to validate \model along two axes: (i) robustness and reliability across diverse operators and motion profiles, and (ii) data utility for training downstream autonomous policies.

\subsection{Robustness and Reliability}

We evaluate robustness through two complementary protocols: real-time teleoperation with diverse human operators, and zero-shot playback of generated motions unseen during training.

\paragraph*{Real-time teleoperation}
We recruit $6$ participants of varying gender and height ($1.47$\,m to $1.94$\,m), the majority of whom are novices with no prior teleoperation experience. Each participant performs a composite loco-manipulation task that simultaneously tests locomotion stability and manipulation precision: the operator teleoperates the humanoid to walk toward a target table, establish a stable stance, and execute a `pick-and-place' maneuver.

As shown in \cref{fig:generalize}, the system maintains consistent stability throughout the task despite substantial anthropometric differences among operators. All novice users achieved successful completion within only $5$--$7$ practice trials, indicating that \model generalizes effectively across diverse body proportions while offering an intuitive, low-barrier interface. We refer the reader to the supplementary video for comprehensive visualizations.

\begin{figure}[t!]
    \centering
    \small
    \includegraphics[width=\linewidth]{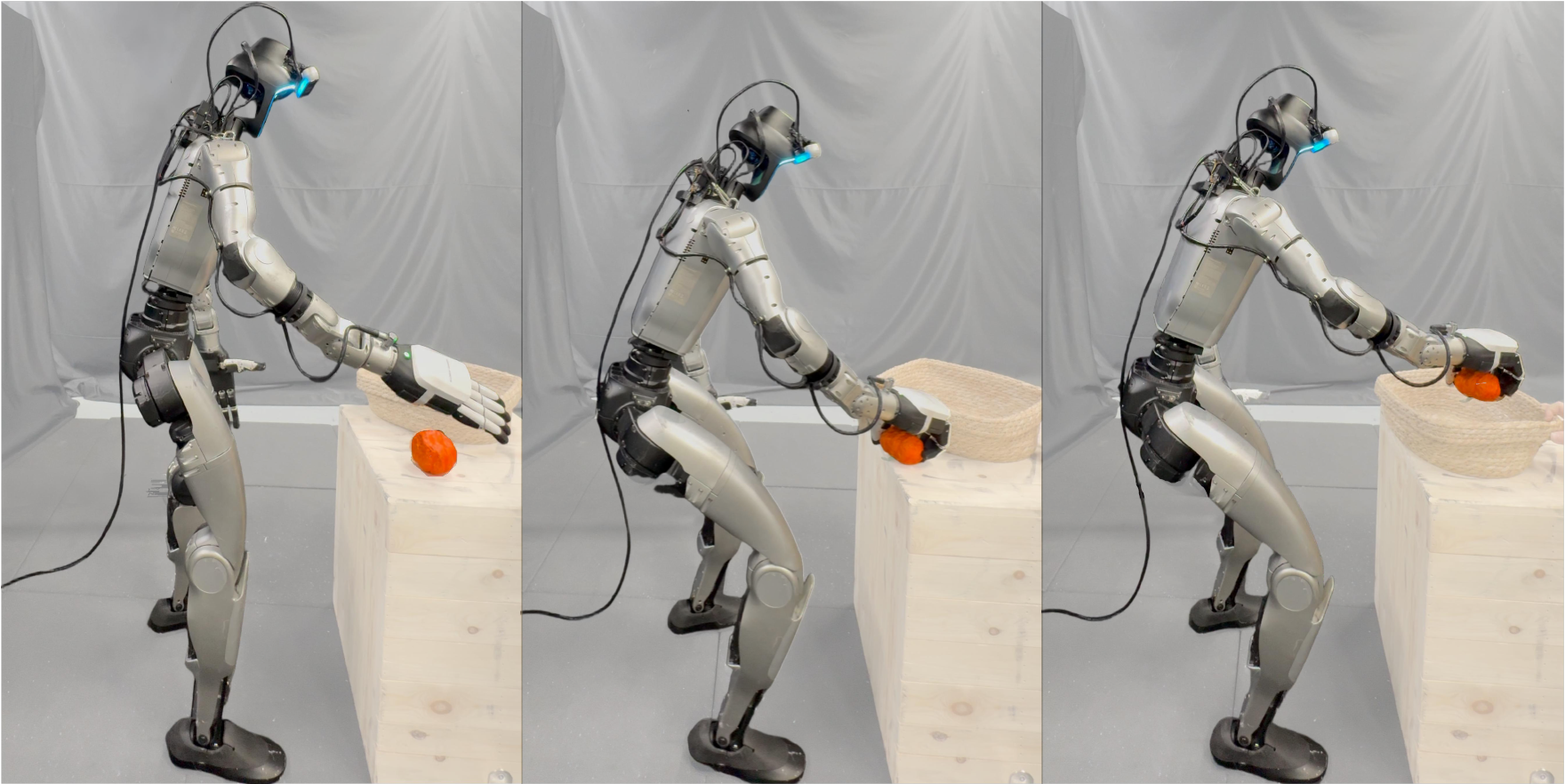}
    \caption{\textbf{Enabling Autonomy via \groot N1.6~\cite{nvidia_gr00t} Post-Training.} We utilize high-fidelity data collected by our system to post-train the \groot N$1.6$~\cite{nvidia_gr00t} foundation model, demonstrating superior full-body control capability.}
    \label{fig:post_train}
\end{figure}

\paragraph*{Zero-shot generalized motion playback}
Since our full-body control policy is trained exclusively on standard \ac{mocap} data, evaluating it on unseen synthetic motions provides a rigorous test of generalization. We use HY-Motion-$1.0$~\cite{hymotion2025} to generate a diverse suite of test motions spanning stationary manipulation, loco-manipulation, and highly dynamic locomotion.

As depicted in the left panels of \cref{fig:teaser} and in \cref{fig:replay}, \model tracks these unseen motion profiles with consistent fidelity, demonstrating robust zero-shot transfer to novel motion sources. This confirms that the system is control-source-agnostic: the same unified policy executes reliably regardless of whether the reference motion originates from a real-time teleoperator or a generative model.

\paragraph*{Qualitative results}
We further visualize real-time teleoperation across a broad range of behaviors on the right side of \cref{fig:teaser} and in \cref{fig:tracking}. The system robustly executes complex, long-horizon maneuvers---`Transporting Coffee Beans' (\cref{fig:teaser}), `Retrieving Tissue from a Low Cabinet' (\cref{fig:tracking}(a)), and `Serving Tea' without spillage (\cref{fig:tracking}(b))---spanning loco-manipulation at varying heights, articulated and deformable object handling, and precise liquid transport. As shown in \cref{fig:tracking}(c), \model also executes highly dynamic motions such as running. Crucially, this full spectrum of behaviors, from precise quasi-static manipulation to dynamic locomotion, is achieved with a single unified policy.

\subsection{Data Utility for Autonomy}\label{sec:vla}

To validate the quality of demonstrations collected via \model, we post-train \groot N1.6~\cite{nvidia_gr00t}, a \ac{sota} \ac{vla} foundation model, on $50$ full-body demonstrations and evaluate it on two manipulation tasks: `Pick-and-Place' (lifting a tomato and placing it into a basket; \cref{fig:tabletop_pick}) and `Squat to Pick-and-Place' (performing the same task from a squatting pose; \cref{fig:post_train}). Notably, \model serves as both the data collection interface and the low-level controller during autonomous inference, imposing stringent demands on the fidelity of the control policy and the stability of the system infrastructure. Implementation details are provided in \cref{sec:app:autonomy}.

As reported in \cref{tab:autonomy}, the autonomous policy achieves \acs{sr} of $85.71\%$ and $80.00\%$ on the two tasks respectively, generalizing to minor variations in object position. These results confirm that \model produces clean, consistent, and kinetically feasible demonstrations, enabling effective downstream policy learning even in low-data regimes.

\begin{table}[t!]
    \centering
    \small
    \setlength{\tabcolsep}{3pt}
    \caption{\textbf{Autonomous policy trained on \model-collected demonstrations achieves high \acs{sr} with only $50$ demonstrations.} We post-train \groot N1.6 on full-body data collected via \model and evaluate on two whole-body manipulation tasks. The results validate that \model produces kinetically feasible, high-fidelity demonstrations suitable for effective downstream policy learning in low-data regimes.}
    \label{tab:autonomy}
    \begin{tabular}{lcc}
        \toprule
        \textbf{Task} & \textbf{Trials} & \textbf{\acs{sr}} \\
        \midrule
        Pick-and-Place & 21 & 85.71\% \\
        Squat to Pick-and-Place & 5 & 80.00\% \\
        \bottomrule
    \end{tabular}
\end{table}

\section{Conclusion}

We have presented \bench and \model, addressing the evaluation and system gaps that hinder practical whole-body humanoid teleoperation. \bench provides the first diagnostic benchmark that disentangles policy performance across stratified motion categories and difficulty levels on unseen motions, replacing coarse aggregate metrics with fine-grained analysis that exposes the narrow specialization of existing systems. Guided by these diagnostics, we identified an optimized training data recipe and demonstrated that the composition of even a small fraction of the training data can decisively steer policy behavior---a finding with broad implications for data curation in humanoid control.

Building on these insights, \model integrates a transformer-based full-body tracking policy with system-level mechanisms---subject-agnostic retargeting and robust communication---into a unified framework that achieves high-fidelity teleoperation on a single consumer GPU with approximately $30$ hours of motion data. The system generalizes across teleoperators of vastly different heights, operates as a control-source-agnostic platform compatible with real-time teleoperation, generated motion playback, and downstream \ac{vla} models, and produces kinetically feasible demonstrations that enable effective autonomous policy learning in low-data regimes.

Looking ahead, several promising directions remain. Scaling the data recipe to close the remaining gap in extreme dynamic regimes (\eg, high jumps) while preserving overall balance is a natural next step. Extending \bench with standardized real-world task protocols would further strengthen its diagnostic utility. Finally, the control-source-agnostic nature of \model opens the possibility of closing the loop between generative motion models and long-horizon autonomous execution in the real world.

\section*{Acknowledgments}
This work is supported in part by the National Key Research and Development Program of China (2025YFE0218200), the National Natural Science Foundation of China (62172043 to W.L., 62376009 to Y.Z.), the PKU-BingJi Joint Laboratory for Artificial Intelligence, the Wuhan Major Scientific and Technological Special Program (2025060902020304), the Hubei Embodied Intelligence Foundation Model Research and Development Program, and the National Comprehensive Experimental Base for Governance of Intelligent Society, Wuhan East Lake High-Tech Development Zone.

\bibliographystyle{plainnat}
\bibliography{reference_header,references}

\clearpage
\crefname{section}{Appendix}{Appendices}    
\Crefname{section}{Appendix}{Appendices}    
\setcounter{table}{0}
\setcounter{figure}{0}
\setcounter{footnote}{0}
\renewcommand{\thefigure}{A.\arabic{figure}}
\renewcommand{\thetable}{A.\arabic{table}}
\renewcommand{\theequation}{A.\arabic{equation}}

\begin{appendices}

\section{Whole-Body Tracking Policy}

\subsection{Problem Formulation}\label{sec:app:prob-formulation}

We formulate whole-body humanoid teleoperation as a motion tracking problem, modeled as a \ac{mdp} defined by the tuple $\mathbb{M} = (\mathbb{S}, \mathbb{A}, \mathbb{T}, \mathbb{R})$, where $\mathbb{S}$ encompasses the full system state including proprioceptive and task-oriented variables, $\mathbb{A} \subset \mathbb{R}^{29}$ represents the target joint angles for the humanoid's actuators, $\mathbb{T}$ denotes the transition dynamics, and $\mathbb{R}$ is the reward function designed to encourage tracking fidelity and physical plausibility. We optimize a policy $\pi$ using the \ac{ppo} algorithm to maximize the expected cumulative reward.

Given a reference target trajectory, the policy minimizes the deviation between the robot's current state and the reference while maintaining balance and stability:
\begin{equation}
    \pi: \mathbb{O} \to \mathbb{A}, \quad \mathbb{A} = \{ \mathbf{a} \mid \mathbf{a} \in \mathbb{R}^{29} \},
\end{equation}
where $\mathbb{O}$ represents the observation space of the policy.

\subsection{Model Structure}\label{sec:app:model_structure}

We employ a Transformer-based Actor-Critic architecture comprising three functional components: a linear input projection, a Transformer backbone, and \ac{mlp} output heads.

The teacher model uses a lightweight Transformer backbone with embedding dimension $d_{\text{model}}=256$ and \ac{ffn} dimension $d_{ff}=512$, processing observations from $4$ input tokens. The student model is designed with higher representational capacity: the embedding dimension is doubled to $d_{\text{model}}=512$ with a corresponding \ac{ffn} dimension of $d_{ff}=1024$, processing observations via $2$ input tokens. Both models employ multi-head attention with $4$ heads.

\subsection{Reward Function and Domain Randomization}

The reward function components and their respective weights are summarized in \cref{tab:reward}. The domain randomization parameters applied during training are detailed in \cref{tab:dr}.

\begin{table}[t!]
    \centering
    \small
    \setlength{\tabcolsep}{3pt}
    \caption{\textbf{Statistical overview of the generated evaluation corpus.} For locomotion tasks, we report the mean velocity alongside the $5$th and $95$th percentiles~(P$5$/P$95$) to exclude transient phases such as startup and stopping transitions. For manipulation tasks, where metrics focus on root or hand height, we report the mean, min, and max values to illustrate the full extent of the reachable workspace. (\textbf{Med}: Medium; \textbf{Loco}: Loco-manipulation; \textbf{Manip}: Manipulation.)}
    \label{tab:eval_stats}
    \raggedright \textbf{(a) Locomotion Speed (m/s)} \vspace{2pt}
    \begin{tabularx}{\linewidth}{L YYY} 
        \toprule
        \textbf{Motion} & \textbf{Min P5} & \textbf{Max P95} & \textbf{Mean} \\
        \midrule
        Walk Slow   & $0.618$ & $1.704$ & $1.026$ \\
        Walk Med    & $0.793$ & $2.193$ & $1.364$ \\
        Walk Fast   & $0.665$ & $2.279$ & $1.584$ \\
        Run Slow    & $0.667$ & $2.883$ & $1.932$ \\
        Run Med     & $0.644$ & $4.151$ & $2.386$ \\
        Run Fast    & $0.685$ & $5.046$ & $2.956$ \\
        \bottomrule
    \end{tabularx}
    \raggedright \textbf{(b) Root Height (m)} \vspace{2pt}
    \begin{tabularx}{\linewidth}{L YYY}
        \toprule
        \textbf{Motion} & \textbf{Min} & \textbf{Max} & \textbf{Mean} \\
        \midrule
        Jump Low    & $0.522$ & $1.206$ & $0.783$ \\
        Jump Med    & $0.466$ & $1.482$ & $0.783$ \\
        Jump High   & $0.351$ & $1.638$ & $0.783$ \\
        Squat Low   & $0.260$ & $0.954$ & $0.610$ \\
        Squat Med   & $0.338$ & $0.831$ & $0.627$ \\
        Squat High  & $0.357$ & $0.818$ & $0.690$ \\
        \bottomrule
    \end{tabularx}
    \raggedright \textbf{(c) Hand Height (m)} \vspace{2pt}
    \begin{tabularx}{\linewidth}{L YYY}
        \toprule
        \textbf{Motion} & \textbf{Min} & \textbf{Max} & \textbf{Mean} \\
        \midrule
        Loco Low         & $0.209$ & $1.040$ & $0.664$ \\
        Loco Med         & $0.628$ & $1.112$ & $0.796$ \\
        Loco High        & $0.626$ & $1.470$ & $0.929$ \\
        Manip Low        & $0.162$ & $1.007$ & $0.698$ \\
        Manip Med        & $0.661$ & $1.103$ & $0.849$ \\
        Manip High       & $0.633$ & $1.387$ & $0.990$ \\
        \bottomrule
    \end{tabularx}
\end{table}

\section{Simulations}

\subsection{Dataset}\label{sec:app:dataset}

\paragraph*{Training dataset}
Our training dataset is partially derived from the TWIST$2$~\cite{ze2025twist2} corpus, where we expand the retargeted motions from $23$ to $29$ \ac{dof} to match our robot's configuration. To align the training data with the insights derived from \bench, we apply heuristic filtering criteria based on root state (position and velocity) and whole-body joint energy to selectively curate the reference motions. Following the data collection protocol established in TWIST$2$, we further augment this base dataset with additional reference motions captured directly via our teleoperation interface. The final dataset is predominantly composed of standing manipulation tasks (approximately $60\%$), totaling $30$ hours of motion data. The distribution of the remaining motion categories is detailed in the data recipe presented in \cref{fig:pipeline}.

\paragraph*{Evaluation dataset}
To generate reference motions that comprehensively reflect the model's capabilities on unseen distributions, we utilize HY-Motion-$1$~\cite{hymotion2025} to synthesize the evaluation dataset forming the core of \bench. Detailed specifications are provided in \cref{tab:eval_stats}. Crucially, this dataset encompasses a diverse array of tasks representative of daily activities, none of which are present in the training distribution. \bench therefore serves as a rigorous testbed for evaluating tracking precision and robustness across multifaceted capabilities.

\begin{table*}[t!]
    \centering
    \small
    \setlength{\tabcolsep}{3pt}
    \begin{threeparttable}
        \caption{\textbf{Reward function configuration.} Summary of reward components and their respective weights used for training. (\textbf{Pos.}: Position; \textbf{Rot.}: Rotation; \textbf{Lin.}: Linear; \textbf{Ang.}: Angular; \textbf{Vel.}: Velocity. The end-effectors consist of the `wrist yaw link' and `ankle roll link.')}
        \label{tab:reward}
        \small
        \begin{tabular}{llc}
            \toprule
            Category & Reward Term & Weight \\
            \midrule
            \textbf{Regularization} 
            & Action Rate Penalty & $-8.0$ \\
            & Contact Air Time Penalty & $-100.0$ \\
            & Joint Acceleration Penalty & $-1.0 \times10^{-7}$ \\
            & Joint Position Limits & $-10.0$ \\
            & Velocity/Action Limits & $-1.0$ \\
            \midrule
            \textbf{Tracking} 
            & Torso Global Pos. / Rot. & $0.5$ \\
            & Full-Body Global Lin. / Ang. Vel. & $1.0$ \\
            & Full-Body Relative Pos. / Rot. & $1.0$ \\
            & End-effector Relative Pos. / Rot. / Lin. / Ang. Vel. & $0.5$ \\
            \bottomrule
        \end{tabular}
    \end{threeparttable}
\end{table*}

\begin{table*}[t!]
    \centering
    \small
    \setlength{\tabcolsep}{3pt}
    \begin{threeparttable}
        \caption{\textbf{Domain randomization.} Specifications of the primary randomization parameters and their respective ranges applied during training. (\textbf{CoM}: Center of Mass; \textbf{RFI}: Random Force Injection; \textbf{nominal}: default values defined in the URDF. $\mathcal{U}(a, b)$ denotes a uniform distribution; $\delta_{x,y,z}$ for the CoM uses specific ranges per axis.)}
        \label{tab:dr}
        \small
        \begin{tabular}{ll>{\raggedright\arraybackslash}p{4cm}}
            \toprule
            Parameter & Sampling Distribution & Details \\
            \midrule
            Action Delay (s) & $\mathcal{U}(0.0, 0.02)$ & --- \\
            Action Noise (rad) & $\mathcal{U}(0.0, 0.02)$ & --- \\
            Link Mass & $[0.9, 1.1] \times \text{nominal}$ & `torso,' `shoulder yaw' \\
            Torso CoM Offset (m) & $\delta_{x} \sim \mathcal{U}(-0.075, 0.075)$, $\delta_{y,z} \sim \mathcal{U}(-0.1, 0.1)$ & --- \\
            Torque RFI & $0.02 \times \text{limit}$ (N$\cdot$m) & --- \\
            Static Friction & $\mathcal{U}(0.3, 2.0)$ & `ankle roll,' `pelvis,' `hip roll,' `knee,' `elbow' \\
            Dynamic Friction & $\mathcal{U}(0.3, 2.0)$ & `ankle roll,' `pelvis,' `hip roll,' `knee,' `elbow' \\
            Stiffness Scale & $[0.95, 1.05] \times \text{nominal}$ & --- \\
            Damping Scale & $[0.95, 1.05] \times \text{nominal}$ & --- \\
            Armature Range & $[0.995, 1.015] \times \text{nominal}$ & --- \\
            \bottomrule
        \end{tabular}
    \end{threeparttable}
\end{table*}

\section{Real-World Experiments}

\subsection{System Setup}

\paragraph*{Hardware setup}
We evaluate \model in real-world scenarios using the Unitree G1-Comp platform~\cite{unitree_robocup}. The robot is equipped with a $2$-\ac{dof} actuated head featuring an Intel RealSense D$455$ depth camera~\cite{intel_realsense_d455}, supporting a range of motion of $\pm 50^{\circ}$ in yaw and $[-90^{\circ}, +22.7^{\circ}]$ in pitch to ensure wide visual coverage for manipulation tasks.

\paragraph*{Deployment architecture}
The full-body control policy executes locally on the robot's onboard computer~(PC$2$). Motion data captured from the teleoperator---using either VR headsets such as PICO~\cite{pico4_ultra} or optical \ac{mocap} systems---is processed and retargeted on an external PC before being transmitted wirelessly to the robot via a router.

\subsection{Data Utility for Autonomy}\label{sec:app:autonomy}

We detail the autonomy task specifications, data collection pipeline, and inference protocols used to validate the fidelity of full-body demonstrations acquired via \model. The complete pipeline for autonomous data collection and task execution will be publicly released to facilitate reproducibility.

\paragraph*{Autonomy task specifications}
We utilize \groot N$1.6$~\cite{nvidia_gr00t}, a widely adopted \ac{sota} \ac{vla} foundation model that maps multimodal observations (natural language instructions and robot states) to executable actions. This setup provides a testbed for evaluating three key aspects: (i) the fidelity of demonstrations collected via our teleoperation pipeline, (ii) the reproducibility of our full-body control policy when functioning as the low-level controller for the \ac{vla}, and (iii) the robustness of the overall system infrastructure in executing the \ac{vla}'s high-level commands.

\paragraph*{Data collection pipeline} 
We provide a whole-body data collection pipeline designed for seamless compatibility with \groot N$1.6$~\cite{nvidia_gr00t}, recording the following data streams:
\begin{itemize}[leftmargin=*,noitemsep,nolistsep,topsep=0pt,partopsep=0pt]
    \item \textbf{Visual observations:} Egocentric RGB streams captured via a head-mounted camera at $30$~Hz, temporally interpolated to $50$~Hz to synchronize with the high-frequency full-body control commands.
    \item \textbf{Proprioceptive states:} Comprehensive state $\mathbf{q}$ encompassing the $29$~\ac{dof} whole-body joint positions, $12$~\ac{dof} Inspire hand states, waist orientation, and the $2$~\ac{dof} head pose.
    \item \textbf{Full-body control commands:} The corresponding command stream $\hat{\mathbf{p}}$ derived from the retargeted results, mirroring the structure of the proprioceptive states to ensure accurate demonstration replay and policy supervision.
\end{itemize}

\begin{figure*}[t!]
    \centering
    \includegraphics[width=\linewidth]{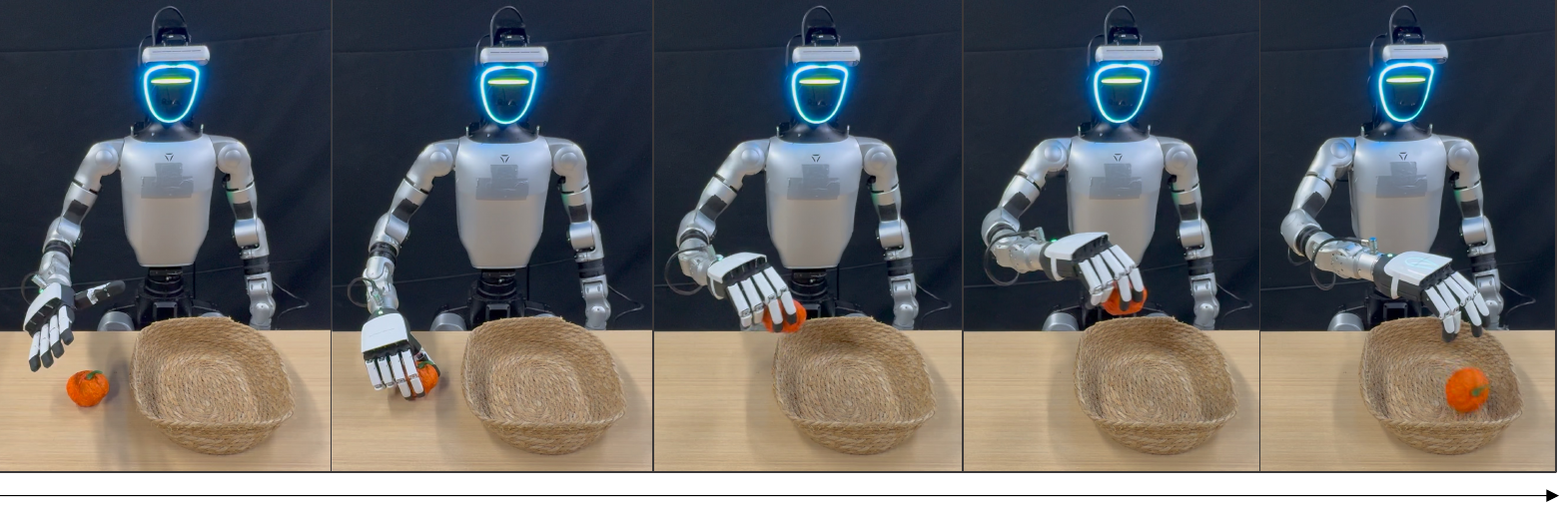}
    \caption{\textbf{Autonomous whole-body pick-and-place tasks executed by \groot N$1.6$ trained on \model-collected demonstrations.} The policy successfully performs full-body pick-and-place from a standing pose, validating the data quality, the fidelity of the control policy, and the stability of the system infrastructure.}
    \label{fig:tabletop_pick}
\end{figure*}

\paragraph*{\ac{vla} model training}
We employ the \groot N$1.6$ \ac{vla} model~\cite{nvidia_gr00t} as the high-level planner and post-train it on our collected dataset following standard protocols. Training is performed on four NVIDIA A$100$ GPUs for approximately $20$k steps with a global batch size of $384$ and a learning rate of $1 \times 10^{-4}$, keeping the backbone vision-language model frozen.

\paragraph*{Inference protocols}
At inference time, \groot N$1.6$ processes the current visual observation, the natural language task description, and the robot's proprioceptive state to predict an action chunk of $16$ steps via a $4$-step denoising process. \model executes the first $8$ actions in an iterative, receding-horizon fashion before the \ac{vla} output is refreshed. The \ac{vla}'s raw joint-space outputs are transformed into body link positions via \ac{fk}, which then serve as observations for \model---a process that inherently introduces potential noise to the control policy.

During autonomous inference, \model functions as the low-level controller while all communication and control command execution share the same system infrastructure as the teleoperation pipeline. This design imposes stringent demands on both the fidelity of the control policy and the stability of the infrastructure. The results in \cref{sec:vla} confirm that \model provides reliable whole-body teleoperation supported by a stable system infrastructure for high-fidelity demonstration collection.

\end{appendices}

\end{document}